\documentclass[sigconf]{acmart}
\AtBeginDocument{%
  }

\copyrightyear{2026}
\acmYear{2026}
\setcopyright{cc}
\setcctype{by}
\acmConference[UIST '26]{The 39th Annual ACM Symposium on User Interface
Software and Technology}{September 28-October 1, 2025}{Busan, Republic of
Korea}
\acmBooktitle{The 38th Annual ACM Symposium on User Interface Software and
Technology (UIST '25), September 28-October 1, 2025, Busan, Republic of Korea}
\acmDOI{}
\acmISBN{}

\usepackage{subcaption}
\usepackage{graphicx}
\usepackage{wrapfig}
\usepackage{xcolor}
\usepackage{listings}
\usepackage[most]{tcolorbox}
\definecolor{codegreen}{rgb}{0,0.6,0}
\definecolor{codegray}{rgb}{0.5,0.5,0.5}
\definecolor{codepurple}{rgb}{0.58,0,0.82}
\definecolor{backcolour}{rgb}{0.95,0.95,0.92}
\newtcblisting{mycode}[1]{
    arc=2pt,
    boxrule=0.5pt,
    colback=backcolour,
    colframe=gray!30,
    listing only,
    listing options={
        language=Python,
        basicstyle=\small\ttfamily,
        keywordstyle=\color{blue},
        commentstyle=\color{codegreen},
        stringstyle=\color{codepurple},
        numbers=left,
        numberstyle=\tiny\color{codegray},
        breaklines=true,
        showstringspaces=false,
        tabsize=4
    },
    title=#1
}

\usepackage[normalem]{ulem}      

\newif\ifstrike
\newif\ifshishi

\newcommand{\strike}[1]{\ifstrike{\color{red}{\texorpdfstring{\sout{#1}}{#1}}}\fi}

\newcommand{\shishi}[1]{\ifshishi{\leavevmode\color{blue}{#1}}\else{#1}\fi}

\begin{document}

\title{Less Is More: Balancing Positive and Negative Space in Visual Concept Blending}

\author{Shishi Xiao}
\affiliation{
  \institution{Brown University}
  \city{Providence}
  \state{Rhode Island}
  \country{USA}
}
\email{shishi_xiao@brown.edu}

\author{Adam J. Coscia}
\affiliation{
  \institution{Georgia Institute of Technology}
  \city{Atlanta}
  \state{Georgia}
  \country{USA}
}
\email{acoscia6@gatech.edu}

\author{David H. Laidlaw}
\affiliation{
  \institution{Brown University}
  \city{Providence}
  \state{Rhode Island}
  \country{USA}
}
\email{dhl@cs.brown.edu}


\begin{abstract}
Graphic designers often blend visual concepts to communicate multiple ideas within a single image, leveraging positive and negative space to create balance, emphasis, and aesthetic appeal.
While computational methods have begun to support automatic concept blending, they largely overlook the role of spatial composition in the design.
To address this gap, we present an automatic pipeline that explicitly applies positive and negative space throughout the blending process.
Our approach first identifies plausible regions for concept integration by combining semantic reasoning from vision–language models with geometric constraints derived from real-world examples. 
Conditioned on these regions, the system generates blended compositions using a hybrid pixel–vector pipeline: diffusion-based inpainting produces a fast, coarse initialization, which is then refined through vector-based optimization at the point level to ensure structural coherence and balanced semantic expression. 
A multimodal agent orchestrates this process as a planner and evaluator, enabling iterative improvement and interpretable control.
Through an evaluation using both baseline comparisons and a user study, we demonstrate greater expressiveness, creativity, and concept recognizability by effectively leveraging positive and negative space. 
We further demonstrate the generalizability of our approach across diverse applications, including controllable image and infographic generation.
\end{abstract}


\begin{teaserfigure}
\centering
    \includegraphics[width=0.86\textwidth]{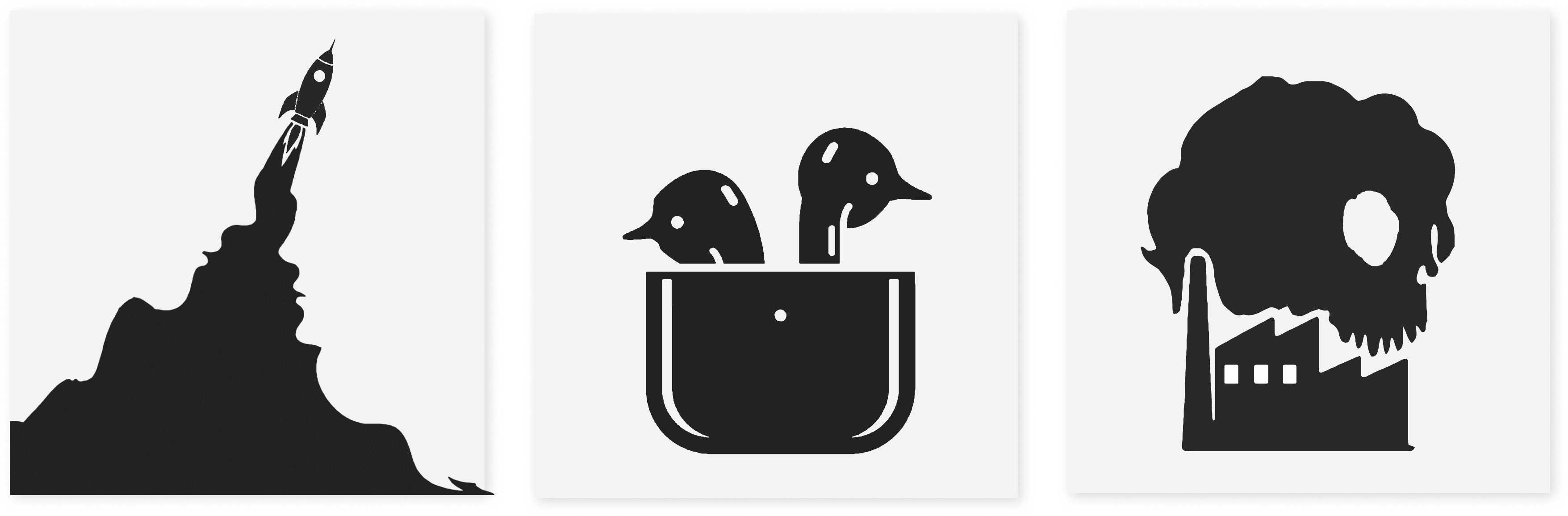}
    \caption{%
        Given two concepts, our method identifies blendable regions and integrates them into recognizable and expressive vector graphics, with user-specified positive or negative blending.
    }%
    \label{fig:teaser}
\end{teaserfigure}


\maketitle

\section{INTRODUCTION}

Visual blending integrates multiple concepts into a single coherent form, conveying dual meanings through a unified representation~\cite{arnheim1954art, ware2019information}. A classical example is the Rubin vase~\cite{rubin1915synsoplevede} (see inset), where a shared contour defines both a vase and two faces, illustrating how foreground and background alternate through positive and negative space. Such compositions extend beyond perceptual illusions and are widely used in graphic design~\cite{lupton2015graphic, lee2007effective}, where designers embed multiple concepts within a shared silhouette, enabling them to share boundaries, compete for visual dominance, and remain recognizable.

\noindent
\begin{minipage}[t]{0.73\linewidth}
  \vspace{0pt}
  Visual blending has been studied in both interactive systems and generative modeling.
Some authoring tools rely on human-crafted heuristics, requiring designers to manually identify compatible regions across concepts~\cite{Chilton:2021:VisiFit, Chilton:2019:VisiBlends}. More recently,
\end{minipage}
\hfill
\begin{minipage}[t]{0.18\linewidth}
  \vspace{-6pt}
  \centering
  \includegraphics[width=\linewidth]{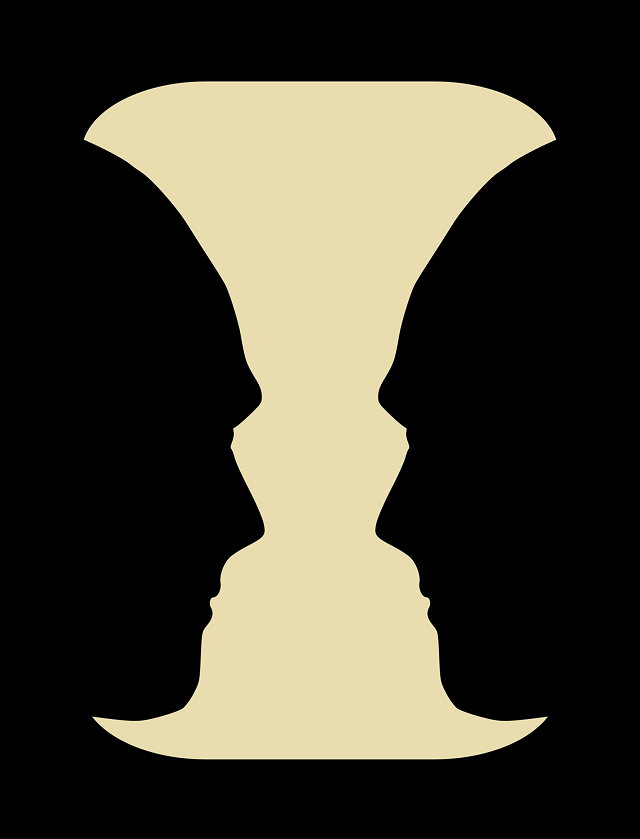}
  \Description{A black-and-white Rubin vase illusion. The central white region appears as a vase, while the surrounding black regions form two human profiles facing each other.}
\end{minipage}
\noindent
diffusion-based generative models have enabled automated blending by leveraging pretrained visual priors, producing diverse forms of blending such as multi-view illusions~\cite{Geng:2025:FactorizedDiffusion} and visual anagrams~\cite{Geng:2024:VisualAnagrams, Burgert:2024:DiffusionIllusions}. Beyond natural images, these techniques expanded blending to support a range of content domains such as text \cite{Iluz:2023:WordAsImage, Xiao:2024:TypeDance}, QR codes \cite{Wu:2024:Text2QR}, and data visualizations \cite{Xiao:2024:ChartSpark, xiao2026chartist, Wu:2023:Viz2Viz}. 

\begin{figure}[!t]
    \centering
    \includegraphics[width=\linewidth]{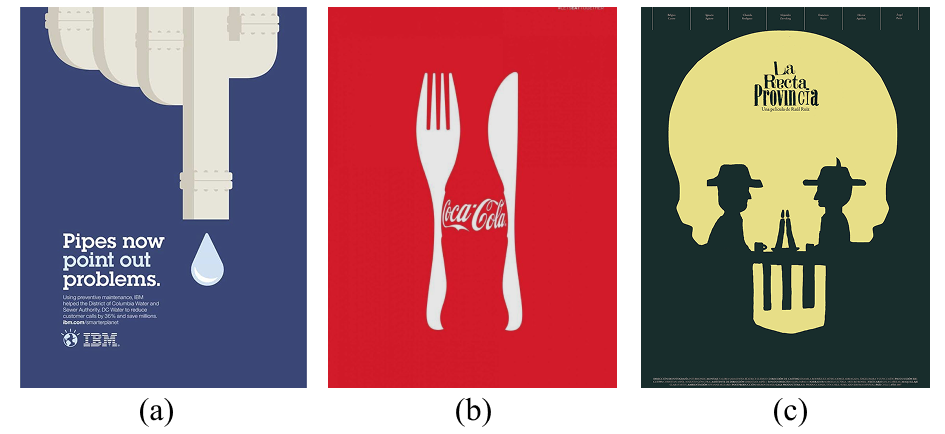}
    \caption{%
        Examples of spatial composition in visual blending: shared-region (a) and positive–negative compositions (b,c).
    }%
    \label{fig:intro_example}
    \Description{%
      Accessibility -- TODO
    }%
\end{figure}

However, despite this progress, the application of space and composition remains underexplored. Prior methods mainly focus on \textit{what} to blend---such as finding object with matched shape or inducing perceptual ambiguity---while paying less attention to \textit{where} and \textit{how} concepts should be composed to balance expressiveness, creativity, and recognizability.
As shown in Figure~\ref{fig:intro_example}, space and composition play a key role in visual blending, shaping the balance and rhythm of a design. Concepts may be integrated within the same foreground region, as in the pipe and hand in Figure~\ref{fig:intro_example}(a), or distributed across foreground and background through positive--negative space, as in the bottle formed by the negative space in Figure~\ref{fig:intro_example}(b) and the human figures embedded within the skull silhouette in Figure~\ref{fig:intro_example}(c).
Inspired by these designs, we focus on blending at the level of shape and contour, abstracting away texture and embellishment to emphasize spatial composition. Under this perspective, we identify two core challenges:

\begin{enumerate}
    \item \textit{Where} to blend, i.e., identifying spatial regions where concepts can be integrated without breaking structural coherence; and
    \item \textit{How} to blend, i.e., constructing a shared contour that preserves recognizability of each concept.
\end{enumerate}

We address these challenges with two key contributions.
First, we derive a set of design guidelines for space composition by analyzing real-world examples through both artistic and geometric lenses. Second, we instantiate these guidelines in a general pipeline that supports automatic concept blending.

Our pipeline consists of two stages that sequentially address the problems of where and how to blend. In the first stage, we identify candidate contour segments using geometric constraints derived from our design principles, and employ a multimodal agent as a planner to refine these candidates with semantic understanding, allowing flexible merging and splitting of segments.
In the second stage, we adopt a hybrid pixel--vector representation to leverage the complementary strengths of both spaces: pixel-based generation enables fast and flexible deformation, while vector representation provides precise point-wise control over contour. We first use an inpainting diffusion model with dynamic mask generation to inject the target concept into the selected contour segments in pixel space. We then convert the result into vector space by optimizing the contour with geometric objectives, including Chamfer loss and bending loss. Next, a multimodal agent evaluates current result from both local and global level, producing a set of concept–score pairs. Guided by these region-specific constraints, we finally optimize the vectorized shape with score distillation loss~\cite{poole2022dreamfusion} to leverage the pretrained prior from diffusion model. Because the final output is represented in vector representation, the results is highly editable and can be seamlessly integrated into real design practice.

We evaluate our framework through quantitative comparison and a human study. Results show that our method improves expressiveness, recognizability, and creativity over existing approaches. We further demonstrate the versatility of our approach across applications, including controllable image and infographic generation.


\section{RELATED WORK}



\subsection{Visual Concept Blending}
\label{sec:related_work_blending}

Visual concept blending originates from studies of ambiguity in image composition, where multiple interpretations coexist within a single representation~\cite{arnheim1954art, rubin1915synsoplevede}.
Classic examples include reversible figures~\cite{Wade:1982:VisualIllusions} and hybrid images that vary with viewing distance~\cite{Oliva:2006:HybridImages}, illustrating how perception can shift under different conditions.

\shishi{%
Prior work offers several related but distinct accounts of how two concepts can be combined into one representation.
Conceptual blending, in the cognitive-science sense~\cite{Fauconnier:1998:ConceptualBlending}, describes how structure from two mental spaces combines to produce \textit{emergent} meaning not present in either input alone. 
Related theories of analogy and conceptual combination~\cite{gentner1983structure, wisniewski1997concepts} instead focus on how structures and features from different concepts are aligned and integrated, without requiring new meaning to emerge.
Our work draws on the latter: we focus on aligning and integrating the spatial structure of two concepts within a shared contour.
}%

Recent advances in deep learning have enabled automated visual blending by incorporating these perceptual principles into image generation.
Diffusion models are particularly effective, encoding multiple signals within a single image using pretrained visual priors~\cite{Ho:2020:DiffusionModels, Yang:2023:DiffusionSurvey}.
Techniques ranging from component-wise conditioning~\cite{Geng:2025:FactorizedDiffusion} to multi-view aggregation~\cite{Geng:2024:VisualAnagrams} and multi-condition optimization~\cite{Burgert:2024:DiffusionIllusions} enable the generation of multi-interpretation images.
However, these methods rely on signal decomposition or global pixel manipulation, offering limited control over spatial composition.
In contrast, we focus on shared contours and spatial organization, where multiple concepts remain simultaneously visible.

To enable shared contour blending, prior work in HCI has explored mapping user intent to visual blending concepts, particularly for metaphor generation and creative ideation.
Systems such as VisiBlends~\cite{Chilton:2019:VisiBlends} and VisiFit~\cite{Chilton:2021:VisiFit} support structured decomposition and recomposition workflows, while tools like MetaMap~\cite{Kang:2021:MetaMap} and CreativeConnect~\cite{Choi:2024:CreativeConnect} enable exploration of conceptual relationships between blending targets.
More recent approaches leverage large language and vision models to automatically resolve the gap between disparate concepts and artistic expression~\cite{Ge:2021:VisConceptBlendingLVMs, Wang:2023:PopBlends, Sun:2025:CreativeBlends, Huang:2025:CreativeSynth, Zhou:2025:ProductMeta}.
While effective for identifying \textit{what} to blend, these systems often struggle to identify \textit{where} and \textit{how} to compose concepts spatially.
Our approach addresses this gap by explicitly modeling region selection to balance positive and negative space design principles.

Visual concept blending has also been applied to creating domain-specific artifacts with structural constraints.
In typography, Word-As-Image~\cite{Iluz:2023:WordAsImage} and TypeDance~\cite{Xiao:2024:TypeDance} deform letterforms to embed semantic meaning while preserving legibility.
In data visualization, systems such as ChartSpark~\cite{Xiao:2024:ChartSpark} and Viz2Viz~\cite{Wu:2023:Viz2Viz} integrate semantic imagery with chart structure to maintain data fidelity.
Functional artifacts like Text2QR~\cite{Wu:2024:Text2QR} embed imagery into QR codes while preserving machine readability.
Rather than developing a task-specific pipeline, we demonstrate the potential of our proposed pipeline to generalize across multiple domains.

\subsection{Computational Approaches for Blending}
\label{sec:related_work_computation}

Automatic visual blending methods are typically pixel- or vector-based, offering complementary strengths in expressiveness and geometric control, which we combine in a hybrid pipeline with pixel initialization and vector refinement.
\subsubsection{Pixel-based Approaches}
Pixel-based methods span from classical image processing to modern generative modeling.
Early techniques combine images via alpha compositing, gradient-domain editing~\cite{perez2023poisson}, and frequency-based fusion~\cite{Oliva:2006:HybridImages}, while later systems introduce perceptual heuristics to match and replace compatible regions across concepts~\cite{Chilton:2019:VisiBlends, Chilton:2021:VisiFit, Sun:2025:CreativeBlends}.
More recently, GANs enable blending through latent interpolation and domain translation~\cite{zhu2017unpaired, park2019semantic}, and diffusion models guide denoising to satisfy multiple constraints, supporting effects such as multi-view illusions, ambiguity, and task-specific blends~\cite{Geng:2024:VisualAnagrams, Geng:2025:FactorizedDiffusion, Burgert:2024:DiffusionIllusions, liao2025diffqrcoder, Huang:2025:CreativeSynth}.
Despite their expressiveness, these methods operate in implicit representations, offering limited control over geometry and spatial composition, and primarily focus on \textit{what} to blend rather than \textit{where} and \textit{how}; we therefore use them to produce an expressive initialization for subsequent geometric refinement using vector-based approaches.

\subsubsection{Vector-based Approaches}
Vector-based methods represent graphics as parametric primitives (e.g., SVG paths), enabling explicit control over shape and composition.
Differentiable rendering frameworks such as DiffVG~\cite{li2020differentiable} allow gradients from perceptual and semantic losses to optimize vector parameters, and recent work incorporates diffusion priors via score distillation sampling (SDS)~\cite{poole2022dreamfusion} to guide generation~\cite{jain2023vectorfusion, xing2023diffsketcher, zhang2024text}.
These approaches support fine-grained control over contours and have been applied to blending tasks in typography, stroke-based illustration, and curve-based abstraction~\cite{Iluz:2023:WordAsImage, liu2025dynamic, cheng2026stroke, berio2025neural}.
However, they often rely on strong initialization and external priors, with optimization that can be slow and unstable, and they do not explicitly address region selection or spatial allocation; we therefore leverage vector representations to refine contours and composition after pixel-based initialization.

\subsection{AI Agents for Design Automation}
\label{sec:related_work_agents}

Agent-based AI systems have emerged that can enable the automation of complex workflows by integrating reasoning, planning, and tool use.
Unlike end-to-end generative models, they decompose tasks into structured subproblems and coordinate components through iterative decision-making.
Foundational approaches such as ReAct \cite{Yao:2023:ReAct} interleave reasoning and action, while frameworks like AutoGPT \cite{Yang:2023:AutoGPT} extend this paradigm to long-horizon, goal-driven planning with subtask decomposition and self-refinement.
These systems demonstrate strong capabilities in multi-step planning and tool coordination \cite{Schick:2023:Toolformer, Valmeekam:2023:PlanBench}, though challenges remain in long-horizon reasoning and reliable execution.

Recent work applies agent-based approaches to creative and visual design, where both semantic intent and visual structure must be considered.
For example, GraphicBench \cite{Ki:2025:GraphicBench} enables LLM-based agents to construct multi-step design workflows by coordinating expert modules and tools, while PosterCopilot \cite{Wei:2025:PosterCopilot} supports layout reasoning and iterative refinement using multimodal models.
However, these systems still face limitations in spatial reasoning, global coordination, and action selection.
We extend this paradigm by incorporating agents as planners and evaluators within a pipeline that explicitly models spatial structure, enabling finer control over where and how visual elements are composed.

\begin{figure*}[!t]
    \centering
    \includegraphics[width=0.92\linewidth]{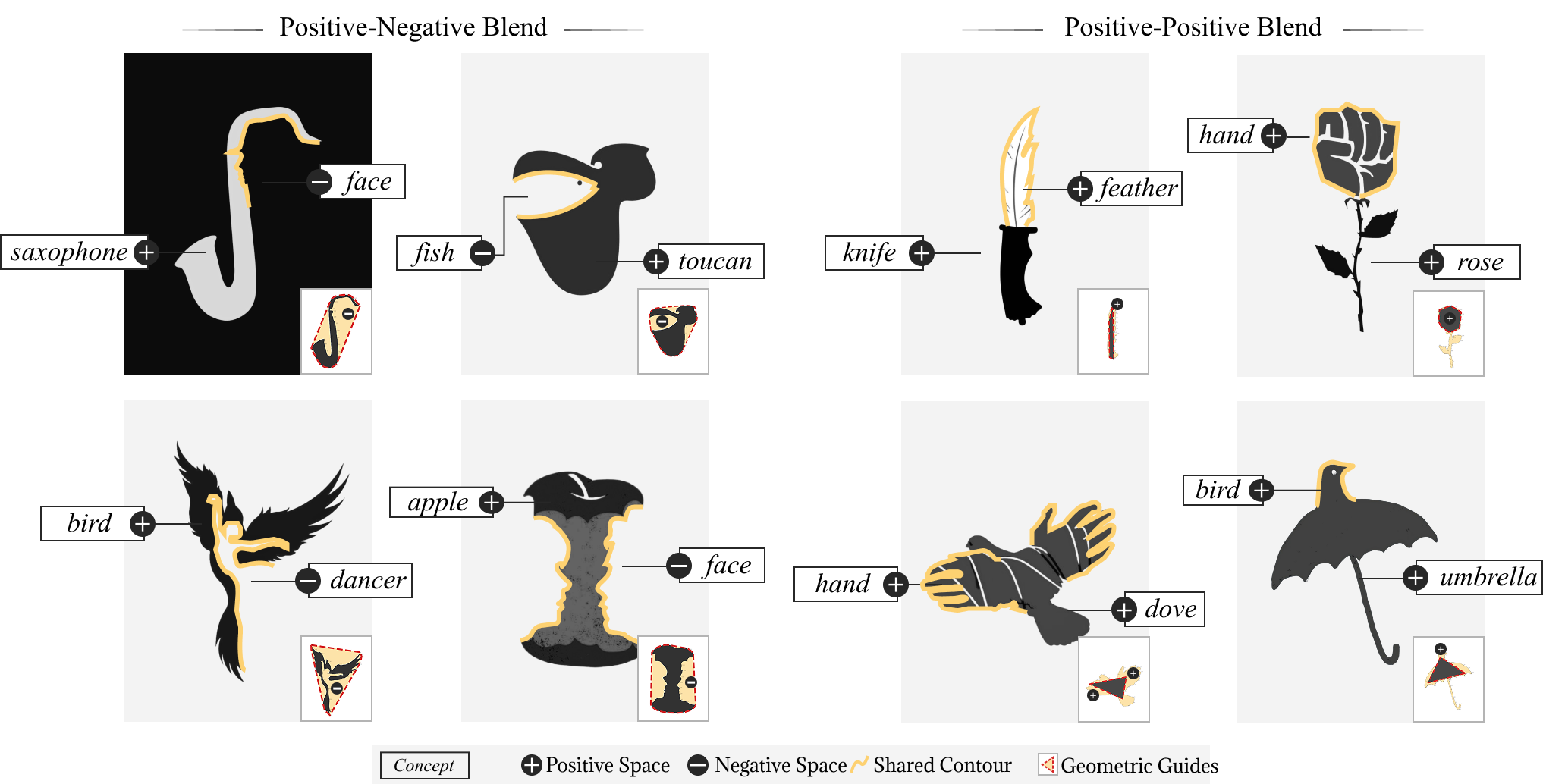}
    \caption{%
        Sampled examples used to explore design guidelines. We analyze two spatial blending modes: positive--negative (left) and positive--positive (right). Each example is annotated with concepts, spatial roles, shared contours, and geometric guides.
    }%
    \label{fig:preliminary_gallery}
\end{figure*}

\section{DESIGN GUIDELINES}
\label{sec:preliminary}


\shishi{To motivate our computational formulation, we consider spatial composition as a central but underexplored dimension of visual blending. We begin by reviewing prior literature on visual perception and graphic design, with a particular focus on positive and negative space~\cite{ware2019information, lupton2015graphic, lidwell2010universal, Kang:2021:MetaMap}. These works suggest three basic design principles: (1) independent interpretations through shared form, (2) negative space as an active element, and (3) contour as a primary driver of perception. However, these principles are typically considered in isolation and formulated for broader design contexts, leaving unclear how they interact to determine the spatial arrangement and shared contours of a visual blend.
}

\shishi{To examine how these principles operate together in practice, we conduct a formative analysis of 100 real-world examples drawn from graphic design, logos, and visual metaphor illustrations. Rather than treating this corpus as an exhaustive taxonomy, we use it to identify recurring relationships among concept roles, positive--negative and positive--positive spatial compositions, and shared contours. We distill these observations into design goals that guide our computational formulation.} 




\subsection{Observations from Real-World Designs}

We analyze a diverse set of examples in which semantic blending is realized at the contour level (Figure~\ref{fig:preliminary_gallery}). 
Across these examples, we annotate three key aspects and derive the following observations:
\begin{itemize}
    \item \textbf{Concepts}: the multiple concepts involved in the blend;
    \item \textbf{Space}: whether concepts are expressed in foreground regions or through negative space;
    \item \textbf{Shared contours}: boundaries that simultaneously support multiple interpretations.
\end{itemize}

\subsubsection{Asymmetric dual interpretation.}
Although blended designs support multiple interpretations, they are typically not perceived equally. 
Instead, designers establish hierarchical emphasis, where one concept dominates attention at first glance, while the secondary concept emerges locally as part of the primary structure. 
This aligns with fundamental principles in graphic design: emphasis and visual hierarchy are used to guide the viewer’s attention and organize elements by importance~\cite{fosco2020predicting}.

\subsubsection{Two modes of spatial composition.}
We observe two common blending modes. 
In \textit{positive--negative blending}, one concept is encoded in the foreground while the other emerges from negative space. 
In \textit{positive--positive blending}, both concepts are expressed within the foreground shape through shared region. Notably, positive--positive blending has been widely discussed in existing computational approaches for visual blending primarily operate within the positive--positive mode while the former one often ignored~\cite{Iluz:2023:WordAsImage, Chilton:2021:VisiFit, Geng:2025:FactorizedDiffusion, liao2025diffqrcoder}.

\subsubsection{Distinct contour-sharing strategies.}

The contour is reused and can be interpreted as dual concepts, but the two blending modes differ in how concepts relate to the contour.
In positive--negative blending, the contour serves as a boundary separating two concepts across space: one concept occupies the foreground, while the other emerges from the negative space on the opposite side. This often results in a convex--concave pairing along the contour.
In contrast, positive--positive blending operate on the same side of the contour, where both concepts reside in the foreground.

\subsection{Design Goals for Visual Blending}


\noindent{\textbf{DG1: Encode dual concepts with clear interpretation hierarchy.}}
Visual blending requires multiple concepts to coexist within a single shape, with visual emphasis guiding perception. 
Effective designs establish an interpretation hierarchy, where a primary concept is immediately recognizable, while secondary concepts are revealed through localized geometric or contour-based cues.

\noindent{\textbf{DG2: Support flexible spatial composition.}}
To support a unified blending formulation, multiple spatial composition modes should be considered, including both positive--negative and positive--positive blending. 
These modes differ in how concepts are spatially arranged, requiring the identification of blending regions that achieve a balance between geometric alignment and semantic interpretability.

\noindent{\textbf{DG3: Enable contour-level optimization.}}
Visual blending is primarily achieved through shape and contour, rather than texture or embellishment. 
As a result, contour-level optimization is essential for adapting and aligning boundaries to encode multiple concepts while maintaining structural coherence and perceptual clarity.

\subsection{From Design Goals to Computational Formulation}
These design goals suggest that visual blending can be formulated along three key aspects, addressing the questions of \textit{what}, \textit{where}, and \textit{how} to blend.


\subsubsection{What to Blend: Concept Role Assignment.}
To realize DG1, we assign asymmetric roles to the two input concepts. The primary concept defines the global silhouette and dominates the overall reading, while the secondary concept is integrated locally along selected contour regions.

\subsubsection{Where to Blend: Region Selection for Blending.}
\label{sec: preliminary_where}

\begin{figure}[h]
    \centering
    \includegraphics[width=\linewidth]{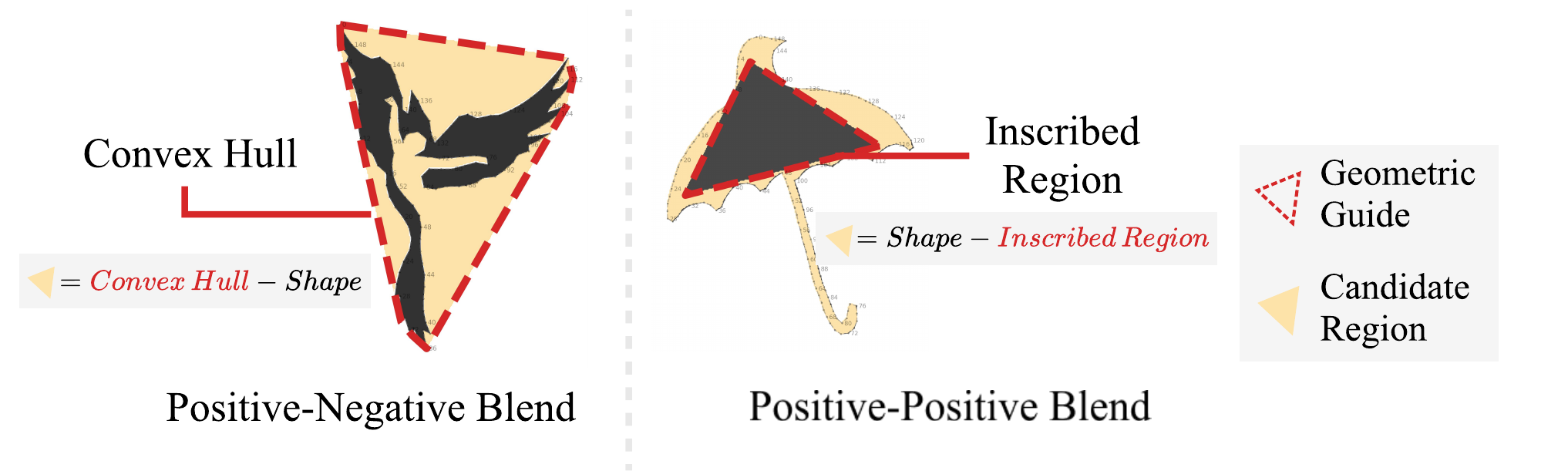}
    \caption{%
        Geometric guides for blending region selection in visual blending. 
        Candidate regions are derived as residuals from the convex hull for positive--negative blending and from an inner region for positive--positive blending.
    }%
    \label{fig:preliminary_where}
\end{figure}

To support flexible spatial composition (DG2), we analyze recurring blending patterns through geometric guides, as illustrated in Figure~\ref{fig:preliminary_where} and the insets in Figure~\ref{fig:preliminary_gallery}. Across examples, candidate blendable regions (highlighted in yellow) consistently occur at locations where the primary shape deviates from its underlying geometric support (indicated by red dashed lines).

\begin{figure*}[t]
    \centering
    \includegraphics[width=\linewidth]{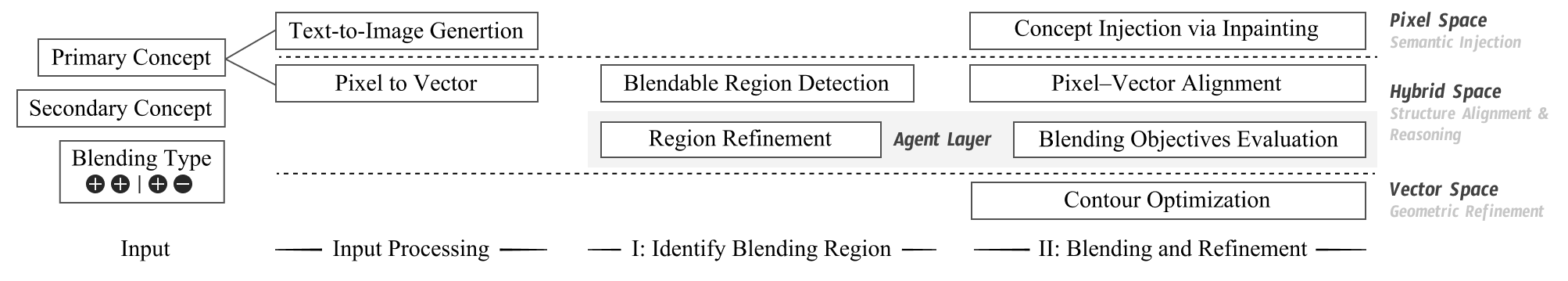}
    \caption{%
        Overview of our pipeline for spatial visual blending. Given a primary and a secondary concept with a specified blending mode, Stage~I identifies blendable regions based on geometric guidance and agent-based planning. Stage~II performs concept integration through a hybrid pixel–vector strategy, including semantic injection via inpainting, geometry-aligned initialization, and contour optimization, producing a coherent blended shape.
    }%
    \label{fig:pipeline}
\end{figure*}

Importantly, the definition of geometric support varies with the blending type. For positive--negative blending, the secondary concept is encoded in negative space, which typically arises from concave regions of the primary shape. We model this by computing the \textit{convex hull} of the primary shape and identifying the interior regions induced by the discrepancy between the shape and its hull.
For positive--positive blending, the secondary concept is integrated into outward protrusions of the primary shape. In this case, we define the geometric support as an \textit{interior support region}, and identify candidate blending regions as the parts of the shape that extend beyond this support.

\subsubsection{How to Blend: Contour-Level Integration.}

To satisfy DG3, we formulate blending as contour-level integration. Selected contour segments are adapted to encode the secondary concept, while the whole structure preserves the global silhouette of the primary concept.
This formulation reflects the fact that different contour segments carry different semantic roles: local segments often correspond to fine-grained components (e.g., a wing for bird), while the overall contour defines the global identity of the shape. As a result, blending requires coordinated local adaptation and global preservation to maintain recognizability of both concepts.

\section{METHOD}
\begin{figure}[h]
    \centering
    \includegraphics[width=\linewidth]{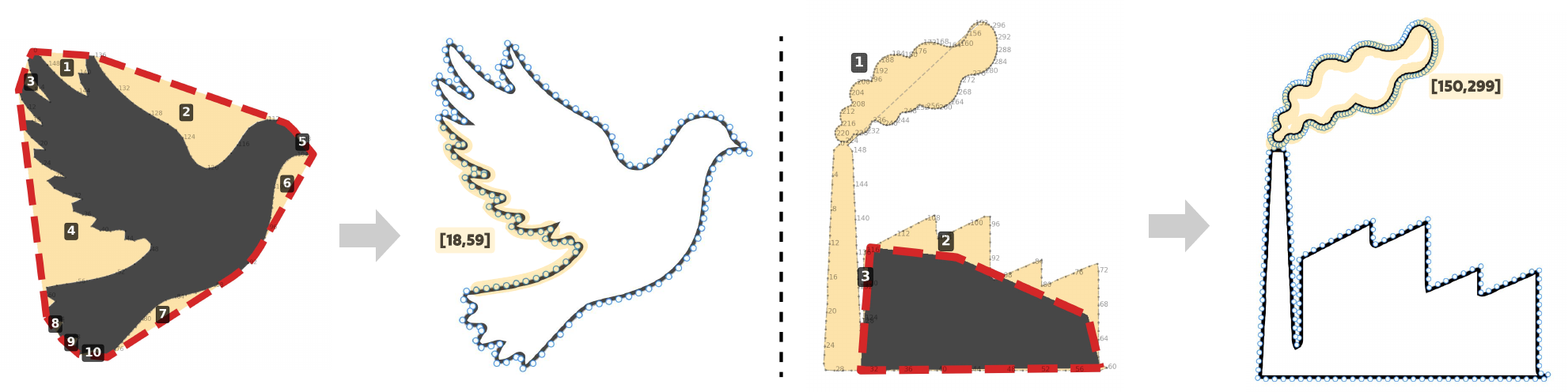}    
    \caption{%
        Stage~I: from candidate regions to a selected contour segment. Geometry-guided detection yields multiple candidate regions, which are consolidated by the planner agent into a specific contour range.
    }%
    \label{fig:method_stage1}
\end{figure}

Given input concepts and a blending mode (positive–positive~\includegraphics[height=0.8em]{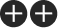} and positive–negative~\includegraphics[height=0.8em]{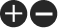}), we aim to generate a new image depicting a coherent silhouette-like form that preserves both concepts \cite{Fauconnier:1998:ConceptualBlending}. 
As shown in Figure~\ref{fig:pipeline}, our framework consists of an input processing stage followed by two stages for \textit{where}- and \textit{how}-to-blend decisions.
\shishi{To illustrate the pipeline under different blending modes, we provide two running examples (Figures~\ref{fig:method_stage1}-\ref{fig:method_output}): \emph{pigeon (+) with human face (-)} and \emph{factory (+) with skull (+)}, corresponding to positive–negative and positive–positive blending, respectively.}

Our pipeline uses the primary concept as the base geometry defining a global silhouette, then integrates the secondary concept into this structure while preserving structural coherence (\textbf{DG1}).
In Stage~I, we identify candidate blending regions using geometric guides (Section~\ref{sec: preliminary_where}), and use a planner agent to select regions based on both geometric constraints and semantic compatibility (\textbf{DG2}).
In Stage~II, given the selected regions, we apply a hybrid pixel–vector approach for blending. We first apply an inpainting model with a dynamically constructed mask to inject the secondary concept, producing a coarse but semantically aligned result. We then convert the result into a vector representation and use an evaluator agent to assess semantic consistency, guiding contour-level refinement (\textbf{DG3}).

\subsection{Instantiating Concepts to Blend}
In our pipeline, users first specify a \emph{primary} concept and a \emph{secondary} concept, along with a user-specified \textit{blending type} (positive–positive or positive–negative).
\shishi{The blend type controls the alignment of the secondary concept to the first.}

We first generate a silhouette of the primary concept using a text-to-image model (\textsc{FLUX.1~DEV}~\cite{blackforestlabs2024flux}). The resulting shape is then converted into an SVG representation using cubic Bézier curves, where each segment is parameterized by one anchor point and two control points denoted by $\mathcal{P} = {\{p_i\}}_{i=1}^N$.
\strike{In our implementation, we empirically set each closed contour to contain 150 control points. We denote the resulting contour as $\mathcal{P} = {\{p_i\}}_{i=1}^N$, where each $p_i \in \mathbb{R}^2$ is ordered along the boundary.}
\shishi{Transforming our initial rasterized image into SVG allows us to (1) automatically detect regions for blending by applying vector calculations; and (2) apply iterative refinement and control over the deformation of the shapes.}

The secondary concept is specified as a text prompt.
Its shape is not predefined; instead, it is implicitly instantiated during blending, where it is injected and progressively aligned to the selected regions of the primary contour under geometric and semantic constraints imposed by the blending type.


\subsection{Stage~I: Identifying Blendable Regions}
Given the SVG representation of the primary concept and the specified blending type, our pipeline then identifies contour segments that are structurally suitable for integrating the secondary concept.
We leverage geometric guides (Section~\ref{sec: preliminary_where}) to computationally instantiate design principles and detect blendable regions along the contour. 
We then employ an AI agent to further refine these candidates with semantic understanding.
We illustrate these two steps in Figure~\ref{fig:method_stage1}.
The output is a set of contour segments.

\subsubsection{Geometry-Guided Blendable Region Detection}
Following Section~\ref{sec: preliminary_where}, we detect blendable regions by measuring geometric discrepancies between the primary shape and its support.
For \textit{positive--negative blending}, we compute the convex hull of the shape and identify contour segments that deviate inward from the hull, corresponding to concave structures that form potential negative space. For \textit{positive--positive blending}, we construct an interior support region and detect outward protrusions relative to this support as candidate regions. Specifically, we estimate the central structure via a distance transform and convert it into a simplified polygon, which is further refined through iterative edge adjustments to remove concave irregularities, yielding a concavity-free support.
Blendable regions are then extracted as contour segments exhibiting significant discrepancies from the corresponding support, as illustrated by the highlighted regions in Figure~\ref{fig:method_stage1}.

\subsubsection{Region Refinement via Planner Agent}
We further refine the candidate regions using a planner agent to determine which contour segments are most suitable for integrating the secondary concept.
\shishi{%
Geometry-based rules can identify multiple plausible blending regions, but they cannot determine which region is most appropriate for the intended concept.
The addition of intelligent LLM-based agents into our pipeline enables us to combine geometric and semantic cues to prioritize blends that allow compound meaning from images to emerge.
For example, an intelligent agent can automatically identify that smoke is more deformable and suitable than a roof for integrating a skull.
This capability enables our outputs to better align with human perceptions of true blending versus juxtaposition of two related concepts.
}%

We employ a vision-language model (\textsc{GPT-5}) as the planner agent, leveraging its semantic understanding and reasoning capabilities. For each candidate region, the agent analyzes both its geometric characteristics and its semantic role within the primary concept, and evaluates its suitability for blending based on several criteria. In particular, The agent prioritizes regions that can be reinterpreted without compromising recognizability (\textit{semantic deformability}), avoid excessive distortion to the overall silhouette (\textit{minimal distortion}), and are sufficiently large to support recognizable integration (\textit{size sufficiency}).
Based on these criteria, the agent refines the candidates by selecting, splitting, or merging regions until only a single contiguous candidate range remains.

\begin{figure}[t]
    \centering
    \includegraphics[width=\linewidth]{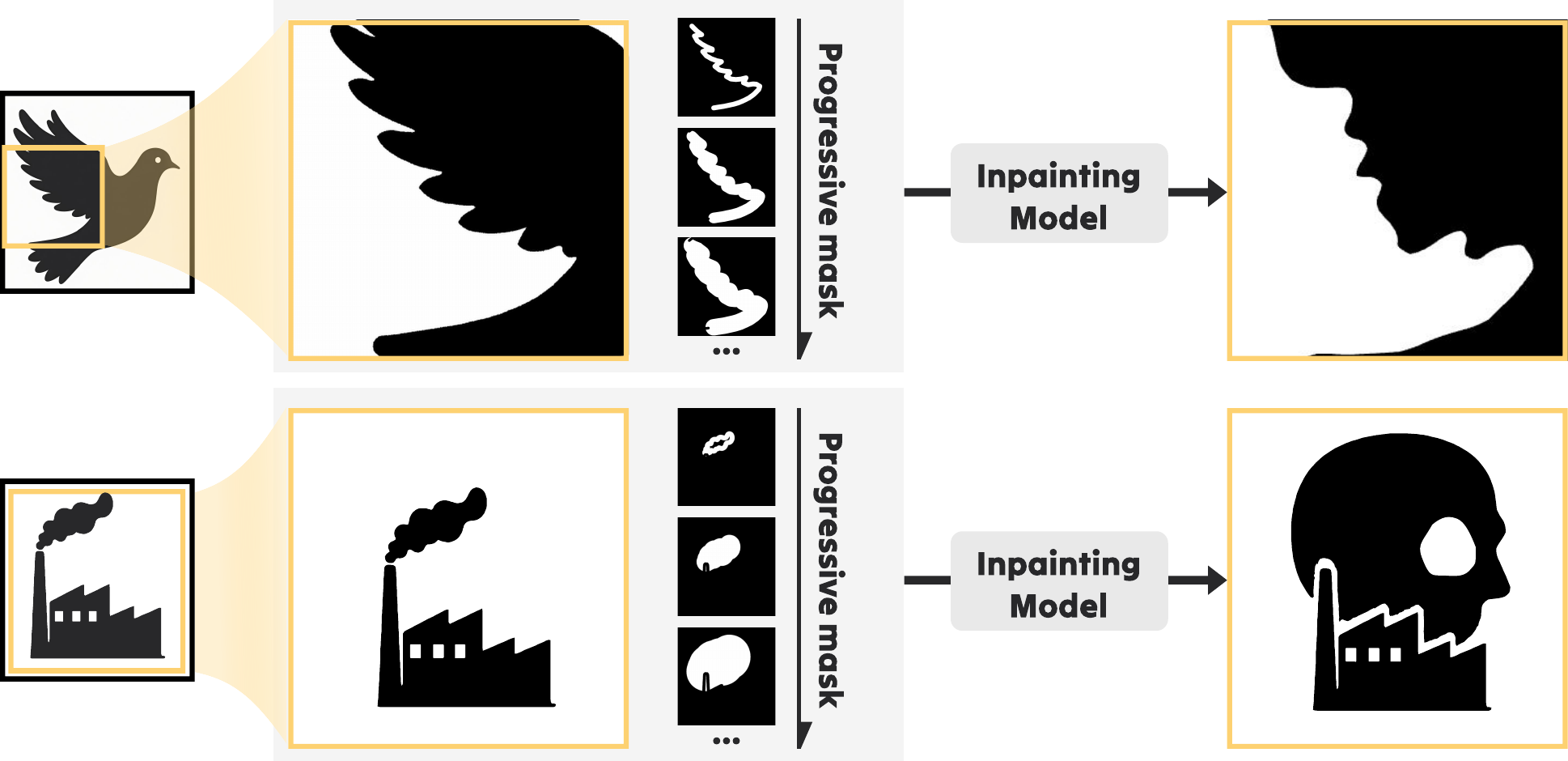}    
    \caption{%
        Stage~II: Inpainting with progressively expanding masks at pixel space.
    }%
    \Description{stage 2 example.}
    \label{fig:method_inpainting}
\end{figure}

\subsection{Stage~II: Blending and Refinement}
Our pipeline then integrates the secondary concept into the selected contour using a hybrid pixel–vector strategy that combines pixel-space generation and vector-level optimization.
\shishi{We describe the intuition and design of each component below; full loss definitions and optimization details are provided in the supplemental material.}

\subsubsection{Concept Injection via Inpainting}
Given the selected contour segment, we perform concept injection in the pixel space using a diffusion-based inpainting model (FLUX.1-dev). 
This step provides a fast semantic initialization by generating the secondary concept conditioned on the primary shape and a dynamically constructed mask, as illustrated in Figure~\ref{fig:method_inpainting}.
\strike{%
The blending mask is constructed by rasterizing $S$ as a polyline.
To provide sufficient space for generation, we progressively expand the mask by increasing the width of polyline across inpainting iterations. 
At the same time, we impose a complementary constraint on the remaining contour by introducing a forbidden mask over non-selected regions. The blending mask is explicitly prevented from overlapping with this forbidden region, ensuring that modifications remain localized while preserving the overall structure of the primary shape.
This stage produces a coarse but semantically aligned injection of the secondary concept, without enforcing overall visual coherence, which are addressed in subsequent steps.
}%
\shishi{%
The mask is built directly from the selected contour segment $S$ and progressively widened across inpainting iterations, giving the model more room to generate the secondary concept while a complementary \emph{forbidden mask} keeps edits from bleeding into the rest of the primary shape.
This stage produces a coarse but semantically aligned injection of the secondary concept; visual coherence with the vector shape is addressed in the steps that follow.
}%

\subsubsection{Pixel–Vector Alignment}
\label{sec:pixel_vector_alignment}
\strike{%
To integrate rasterized details into a vector representation, we perform a constrained optimization over the selected segment $S$, rather than re-vectorizing the entire shape.
A global re-vectorization would introduce a new set of vertices and break the established indexing of $\mathcal{P}$.
}%
\shishi{%
At this stage, the inpainted result lives in pixel space and needs to be brought back into our vector representation.
Rather than re-vectorizing the whole shape -- which would generate a new set of vertices and break the indexing of $\mathcal{P}$ established in Stage~I -- we locally re-fit only the selected segment $S$ to match the inpainted region, while keeping the rest of the contour untouched.
}%
\shishi{Concretely, we extract the target contour from the inpainted region as a point cloud and move the Bézier control points of $S$ toward it, subject to a smoothness constraint that discourages sharp turns and self-intersections.}
\strike{%
We extract the target contour from the inpainted region as a point cloud $\mathcal{T}$, and rearrange the Bézier control points within $S$ by solving the following optimization problem
\begin{equation}
\min_{S} \ \mathcal{L}_{\text{chamfer}}(S, \mathcal{T}) + \lambda \mathcal{L}_{\text{bending}}(S).
\end{equation}
The Chamfer loss aligns the vector contour with the target geometry \(\mathcal{L}_{\text{chamfer}}(S, \mathcal{T}) = \tfrac{1}{|S|} \sum_{p_i \in S} \min_{q_j \in \mathcal{T}} \|p_i - q_j\|_2^2.\) To ensure structural coherence, we introduce a bending loss that regularizes local curvature:
\(\mathcal{L}_{\text{bending}}(S) = \sum_{p_i \in S} \left(1 - \cos(\theta_i)\right)^2\), where $\theta_i$ is the angle between $(p_i - p_{i-1})$ and $(p_{i+1} - p_i)$. This term penalizes sharp turns and prevents self-intersections or jagged artifacts during deformation.
}%
\shishi{The corresponding Chamfer and bending loss terms are given in the supplemental material.}
This yields an updated SVG contour that is geometrically aligned while preserving the original parameterization of $\mathcal{P}$, giving a coherent initialization for the refinement step below.

\subsubsection{Blending Objectives Evaluation}
\begin{figure}[t]
    \centering
    \begin{subfigure}{\linewidth}
        \centering
        \includegraphics[width=\linewidth]{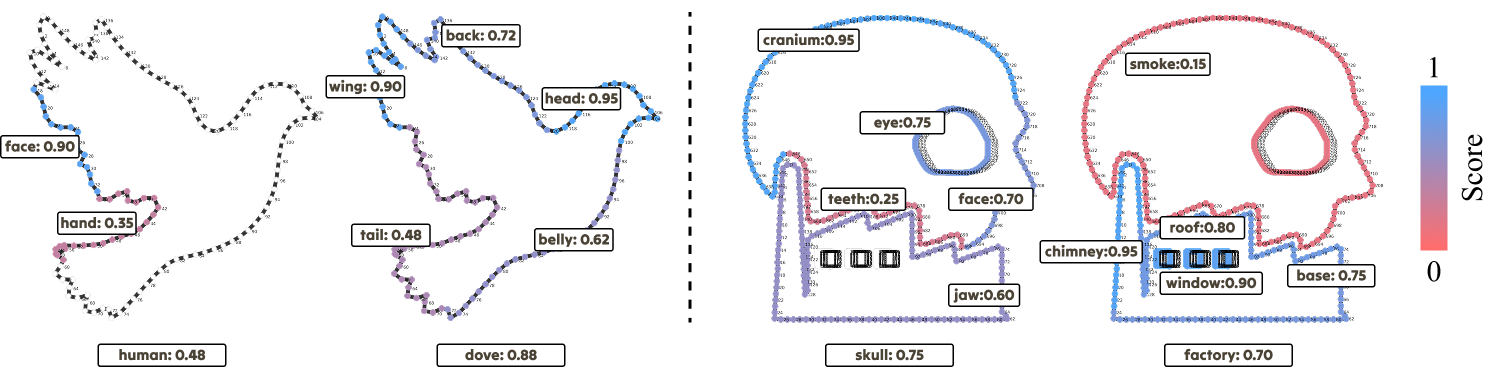}
        \caption{Concept-level expressiveness evaluation.}
        \label{fig:method_eva}
    \end{subfigure}

    \begin{subfigure}{\linewidth}
        \centering
        \includegraphics[width=0.92\linewidth]{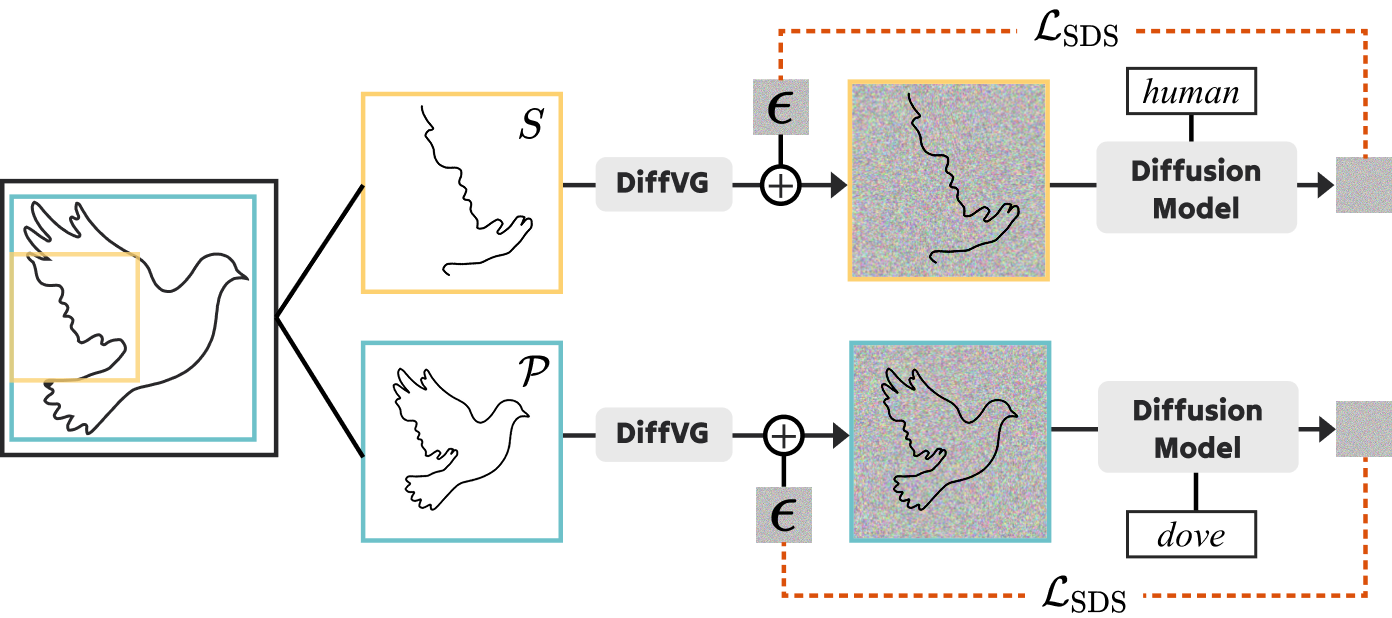}
        \caption{Diffusion-guided contour optimization.}
        \label{fig:method_sds}
    \end{subfigure}
    \vspace{-2em}
    \caption{Stage II: Evaluation and optimization.}
    \label{fig:method_stage2}
\end{figure}

Before further contour optimization, we use an evaluator agent to assess how well the current blend expresses each concept, so refinement can be targeted rather than applied uniformly.
As shown in Figure~\ref{fig:method_eva}, evaluation happens at two levels: \emph{globally}, over the blended shape as a whole, and \emph{locally}, over the specific segments associated with each semantic component.
\strike{%
Formally, we represent the evaluation as a set of concept--score pairs
\(
\mathcal{E} = \{(c_k, s_k)\}_{k=1}^K,  s_k \in [0,1],
\)
where each $c_k$ denotes a semantic component of the blended concept and $s_k$ measures its expressiveness.
}%
For example, in the factory case, the concept is decomposed into components such as smoke, chimney, and building base, each scored independently and each tied to its own segment of the contour.
\strike{%
At the local level, each component $c_k$ is associated with a contour segment $S_k \subseteq S$, enabling spatially grounded evaluation along the curve. Regions with lower scores indicate insufficient or ambiguous expression of the intended concept.
We use these scores to guide optimization by prioritizing low-expressiveness regions.
In practice, we apply a threshold $\tau = 0.5$ and focus refinement on segments where $s_k < \tau$.
}%
\shishi{Components that score below a threshold ($\tau=0.5$ in practice) are flagged as under-expressed and prioritized in the optimization stage that follows.}



\subsubsection{Contour Optimization}
As shown in Figure~\ref{fig:method_sds}, we refine the vector path using differentiable rasterization (DiffVG~\cite{li2020differentiable}) together with Score Distillation Sampling (SDS), where a diffusion model acts as a semantic prior that pulls the contour toward the target concepts.
\strike{%
The SDS objective is defined as
\begin{equation}
\mathcal{L}_{\text{SDS}} = \sum_{k\in [K]} (1 - s_k) \cdot 
\mathbb{E}_{t,\epsilon} \left[
\left\| \epsilon_\theta(x_t; c_k) - \epsilon \right\|_2^2
\right],
\end{equation}
where $x_t$ is the noised rendering of the current contour at timestep $t$, and $\epsilon_\theta$ denotes the noise prediction conditioned on prompt $c_k$.
The loss is weighted by $(1 - s_k)$, prioritizing under-expressed concepts. We jointly optimize it with $\mathcal{L}_{\text{bending}}$ to avoid contour intersections.
}%
\shishi{%
Under-expressed components (per the evaluator's scores) are weighted more heavily in this process, and the same smoothness constraint from Pixel–Vector Alignment (Section~\ref{sec:pixel_vector_alignment}) is applied jointly to prevent self-intersections.
The full SDS objective is given in the supplemental material.
}%

This evaluate-then-optimize loop supports both global and local refinement (\textbf{DG3}): globally, it aligns $S$ and $\mathcal{P}$ with the high-level concepts (Figure~\ref{fig:method_sds}); locally, since a semantic component may correspond to overlapping segments across the two concepts, we optimize those overlapping segments jointly so the blend stays consistent at their interface.

\subsubsection{Output and Editability}
\begin{figure}[h]
    \centering
    \includegraphics[width=0.93\linewidth]{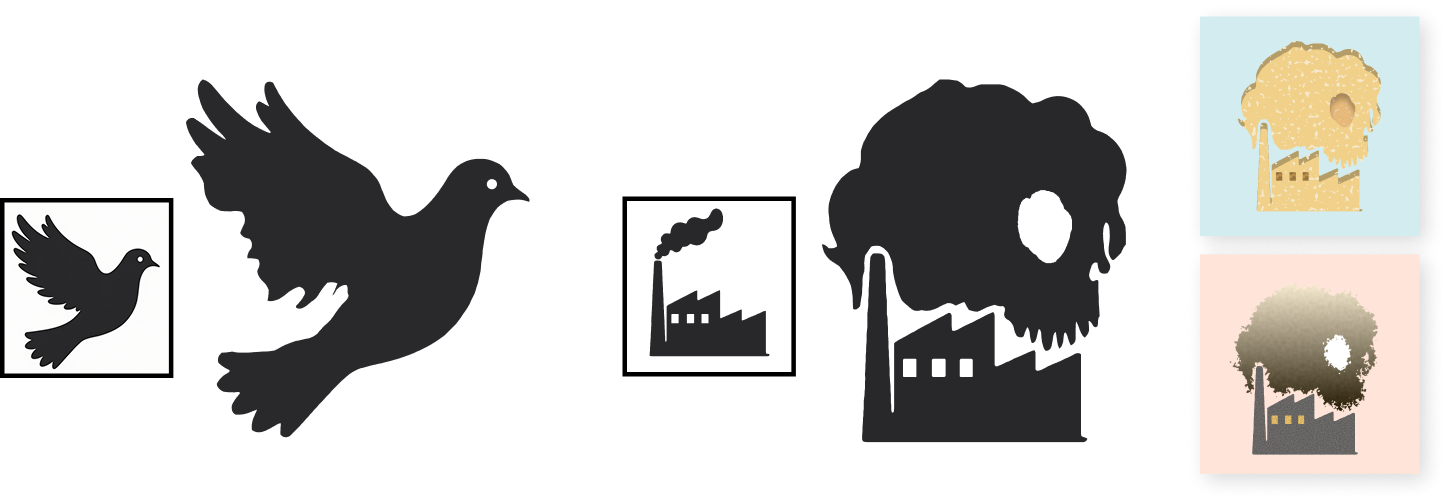}  \caption{Example outputs of visual blending.}
    \label{fig:method_output}
\end{figure}

Example results are shown in Figure~\ref{fig:method_output}. We also demonstrate stylized variants by modifying colors and applying textures and filters. As the outputs are in vector format, they remain fully editable, enabling seamless integration into downstream design workflows.

\section{Evaluation}

\begin{figure*}[t]
    \centering
    \includegraphics[width=0.92\linewidth]{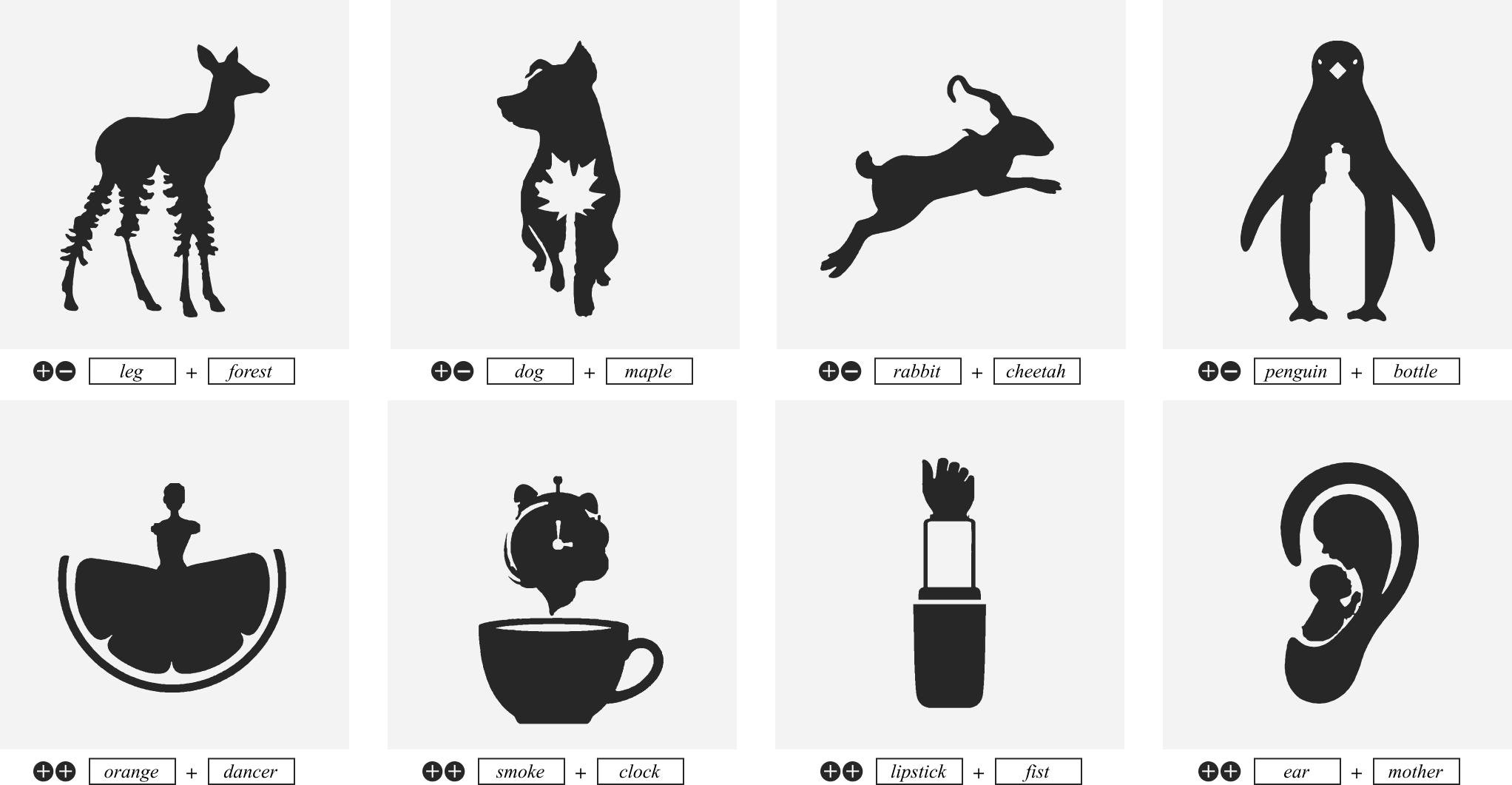}    
    \caption{%
        Result gallery of visual concept blending, including both positive–negative~\includegraphics[height=0.8em]{figs/icon_pn.png} and positive–positive blending~\includegraphics[height=0.8em]{figs/icon_pp.png}.
    }%
    \label{fig:eva_galerry}
    \Description{%
      Accessibility -- TODO
    }%
\end{figure*}

To explore diverse blending scenarios, we use ChatGPT to generate input pairs, including primary and secondary concepts, and corresponding blending types.
We present some of the generated results in Figure~\ref{fig:teaser} and Figure~\ref{fig:eva_galerry}. 
With these materials, we then evaluate our approach through both a quantitative comparison against advanced commercial models (Section~\ref{sec:baseline_comparison}) as well as through a user study with novice participants (Section~\ref{sec:user_study}).
To further examine the generalizability of our pipeline, we finally explore the application of visual blending across multiple application scenarios (Section~\ref{sec:case_study}).
\shishi{Across these studies, our goal is to assess how well the generated results support human interpretation of the intended blend, rather than to optimize a single automated metric in isolation.}

\subsection{Baseline Comparison}
\label{sec:baseline_comparison}

As no existing pipeline is specifically designed for compositional visual blending, we first compare our method with advanced commercial models, including both pixel-based and vector-based approaches in a zero-shot setting.

For pixel-based methods, both concepts are provided as text prompts; for vector-based methods, we additionally supply the SVG of the primary concept for structure-aware generation. All methods are evaluated on 20 samples.
We use three metrics for quantitative evaluation: (1) \textbf{CLIP-T} for recognizability, computed as the average text–image similarity for both concepts~\cite{radford2021learning}; (2) \textbf{CLIP-IQA} for perceptual quality~\cite{wang2022exploring}; and (3) \textbf{Alignment}, measuring adherence to the specified blending type.

\begin{figure}[!t]
    \centering
    \includegraphics[width=\linewidth]{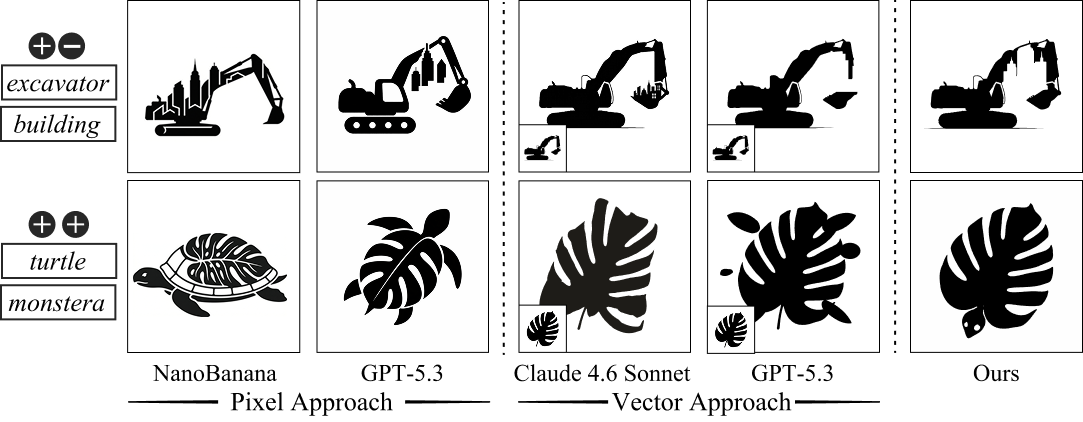} 
        
    \caption{%
        Qualitative results of baseline comparison.
    }%
    \label{fig:eva_baseline}
    \Description{%
      Accessibility -- TODO
    }%
\end{figure}

\subsubsection{Result Analysis}

Results are presented in Figure~\ref{fig:eva_baseline} and Table~\ref{tab:eva_baseline}.
\shishi{Rather than optimizing any single metric, our goal is to balance recognizability, perceptual quality, and compositional alignment, since a successful blend must remain identifiable while still reading as a unified structure rather than a simple juxtaposition.}

Our method achieves the best performance in compositional alignment, outperforming the second-best baseline by 39.4\% (0.902 vs. 0.647), indicating stronger control over the intended blending type. 
In terms of perceptual quality and recognizability, our method ranks third and second, respectively, remaining close to the top baselines. We observe that pixel-based methods often achieve higher scores in these metrics but perform worse in compositional alignment, as they typically rely on spatial placement rather than unified structural integration.
In contrast, our approach encodes the secondary concept through negative space and contour deformation, which may be less favored by CLIP-based metrics.

\shishi{%
Figure~\ref{fig:eva_baseline} highlights the compositional differences between methods. Pixel-based approaches often rely on overlay or insertion, while existing vector methods may leave disconnected components or let one concept dominate. In contrast, our method realizes the requested blending mode:  the city is carved into the excavator as meaningful negative space while preserving the excavator's overall silhouette, and the turtle and leaf share a unified contour. This produces stronger structural integration, consistent with our higher alignment score.
}%

\shishi{%
To better understand this qualitative advantage, we next conduct human evaluation for a more comprehensive assessment.
Our goal was to complement these automatic metrics with human evaluation of the same baselines in the following section (Table~\ref{tab:user_study}).
}%

\begin{table}[!t]
\centering
\small
\setlength{\tabcolsep}{5pt}
\renewcommand{\arraystretch}{1.1}
\begin{tabular}{l|ccc}
\toprule
Method 
& CLIP-T $\uparrow$ 
& CLIP-IQA $\uparrow$ 
& Alignment $\uparrow$ \\
\midrule
NanoBanana~\cite{google2026nanobanana} & \textbf{0.235}  & \underline{0.636} & 0.412 \\
GPT-5.3 (Pixel)~\cite{openai2026gpt53} & 0.189 & \textbf{0.645} & \underline{0.647} \\
Claude Sonnet 4.6~\cite{anthropic2026claude46} & 0.166 & 0.597 & 0.510 \\
GPT-5.3 (Vector)~\cite{openai2026gpt53} & 0.157  & 0.612 & 0.633 \\
\midrule
\textbf{Ours} & \underline{0.193} & 0.629 & \textbf{0.902} \\
\bottomrule
\end{tabular}
    \vspace{0.5em}
\caption{Quantitative comparison results. Best results are in bold and second-best are underlined.}
\label{tab:eva_baseline}
\vspace{-1.5em}
\end{table}

\subsection{User Study}
\label{sec:user_study}

\begin{figure*}[!t]
    \centering
    \includegraphics[width=0.85\linewidth]{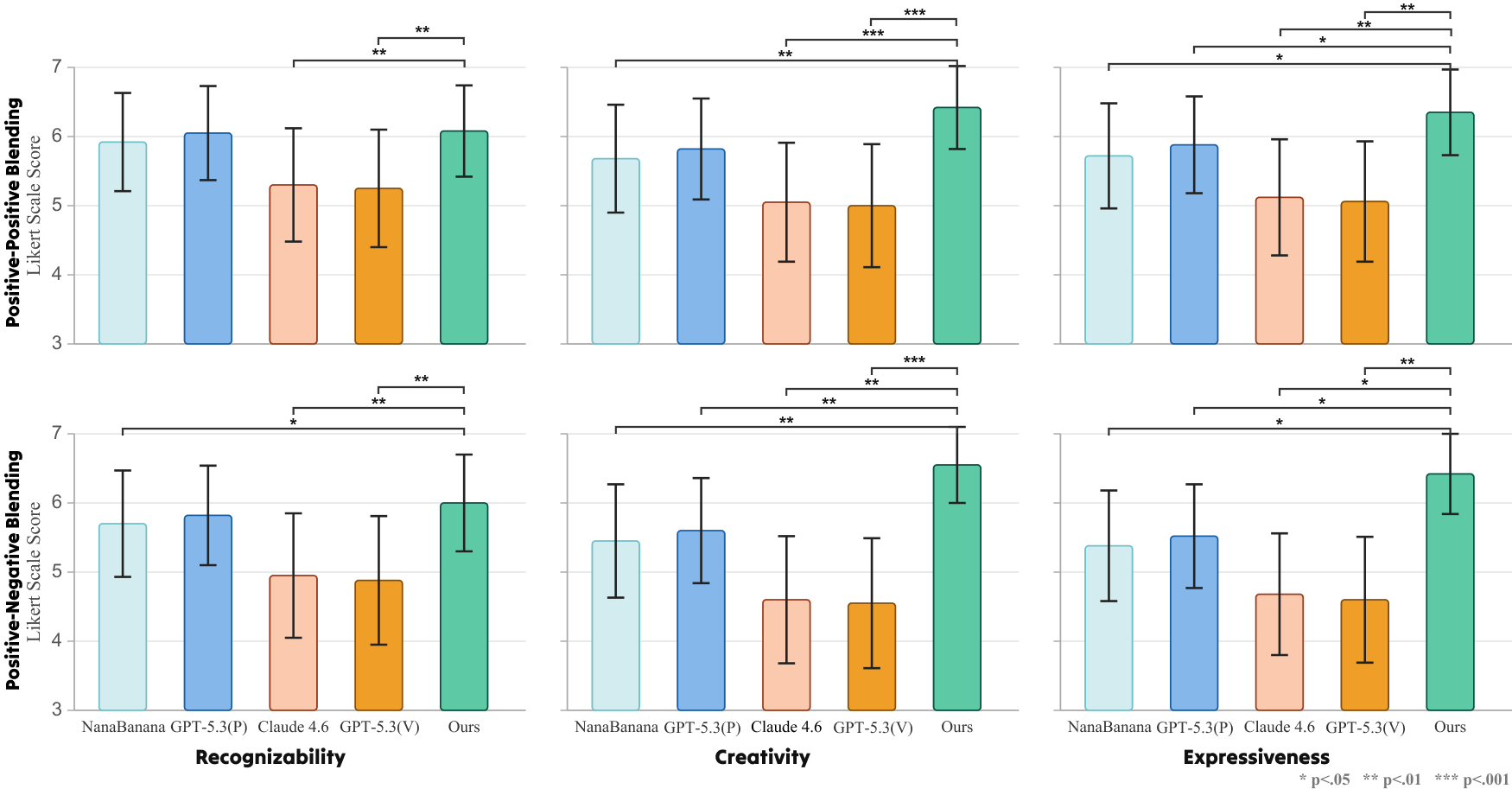}    
    \caption{Result of 7-point Likert scale ratings of Recognizability, Creativity and Expressiveness. Horizontal brackets indicate pairwise significant difference, and the error bars show standard error.
    }%
    \label{fig:eva_user}
\end{figure*}

\shishi{%
To complement the automatic metrics above with audience-level interpretation, we next ran a user study ($n=12$) to examine whether the generated blends communicate the intended concepts, blending regions, and blending modes to non-expert viewers.

We make an important trade-off in our study design -- novice participants were recruited as representative evaluators of audience perception rather than as proxies for expert designers, since successful visual blends must ultimately communicate to general viewers regardless of who creates them.
To provide more control over the process, we provide study participants with an interface that exposes several parameters to control the visual blending process: pipeline design decisions, e.g. geometry-guided candidate regions and semantic reasoning for where to blend (Figure~\ref{fig:method_stage1}), and concept-level scores for different contour segments (Figure~\ref{fig:method_stage2}a).
}%


\subsubsection{Setup.}
We recruited 12 novice participants. All participants had prior exposure to AI-generated images and had seen visual blending examples before, but none had formal training in graphic design or visual design.
We adopted a within-subject design and randomized the presentation order to reduce ordering effects and participant bias.

The study consists of three tasks: a \textit{Where} task, a \textit{How} task, and a subjective questionnaire.
Each task contains 10 image samples, with no overlap of samples across tasks. Positive and negative blending cases are evenly distributed in each task.
Since the \textit{Where} task requires a reference shape of the primary concept for users to select blendable regions, it is only applicable to vector-based methods that start with an explicit primary shape.
This task focuses on evaluating the spatial decision of where the secondary concept is integrated during the design process.
The \textit{How} task and the questionnaire evaluate the perceptual quality of the final results, and include both pixel-based and vector-based baselines.

\begin{table}[!t]
\centering
\small
\setlength{\tabcolsep}{5pt}
\renewcommand{\arraystretch}{1.1}
\begin{tabular}{l|c|cc|cc}
\toprule
\multicolumn{1}{c|}{Method}
& IoU $\uparrow$ 
& \multicolumn{2}{c|}{Accuracy $\uparrow$} 
& \multicolumn{2}{c}{Time (s) $\downarrow$} \\
&  & \includegraphics[height=0.8em]{figs/icon_pp.png} & \includegraphics[height=0.8em]{figs/icon_pn.png} & \includegraphics[height=0.8em]{figs/icon_pp.png} & \includegraphics[height=0.8em]{figs/icon_pn.png} \\
\midrule
NanoBanana 
& -- 
& \underline{0.900} & 0.850 
& \textbf{3.8} & \textbf{4.1} \\

GPT-5.3 (Pixel) 
& -- 
& 0.894 & \underline{0.867} 
& 4.1 & 4.9 \\

Claude 4.6 
& \underline{0.41}
& 0.667 & 0.617 
& 4.2 & 5.3 \\

GPT-5.3 (Vector) 
& 0.36
& 0.650 & 0.533 
& \underline{3.9} & 4.6 \\

\midrule
\textbf{Ours} 
& \textbf{0.58} 
& \textbf{0.917} & \textbf{0.893} 
& \underline{3.9} & \underline{4.2} \\
\bottomrule
\end{tabular}
\vspace{0.5em}
\caption{User study results of Task 1 and Task 2.}
\label{tab:user_study}
\vspace{-1.5em}
\end{table}

\subsubsection{Procedure.}
Participants first completed a brief tutorial with one practice example. The study then included three tasks.

\medskip
\noindent\textbf{Task 1: Blendable region identification (\textit{15 mins}).}
This task evaluates whether the blending regions selected by models align with users' design choices. 
In each trial, participants were shown the primary concept in SVG format, the text description of the secondary concept, and a specified blending type.
\shishi{They first used a lasso tool over the SVG contour points to select the region they considered most suitable for integration.
After making their own selection, they reviewed the model's intermediate outputs, including geometry-based candidate regions, semantic reasoning, and segment-level scores, and compared these with the model's final region selection.}
We compute spatial agreement using Intersection-over-Union (IoU) between the user-selected region \(U \subseteq P\) and model-selected region \(M \subseteq P\): 
\(\mathrm{IoU}(U, M) = \frac{|U \cap M|}{|U \cup M|}\).

\medskip
\noindent\textbf{Task 2: Dual concept recognition (\textit{15 mins}).}
This task evaluates how well users can recognize the two concepts from the blending result.
Participants viewed generated results and identified the perceived primary and secondary concepts. We measure recognition accuracy and response time.

\medskip
\noindent\textbf{Task 3: Questionnaire and interviews (\textit{20 mins}). }
Participants then rated results on a 7-point Likert scale along three dimensions:
(1) \textit{Recognizability}: how easy it is to recognize the concepts in the image;
(2) \textit{Creativity}: how creative the chosen blending location and composition are;
(3) \textit{Expressiveness}: how clearly and aesthetically both concepts are visually expressed.

\subsubsection{Result Analysis.}
\label{sec:user_study_results}







\shishi{We first examine the quantitative measures from Tasks 1–2 (region agreement and recognition performance) before turning to thematic analysis of participants' qualitative interpretations in Task 3.}

As shown in Table~\ref{tab:user_study}, our method achieves the highest IoU, outperforming vector-based baselines by 41.5\% and 61.1\% relative improvement, indicating that the blending regions selected by our method are more consistent with users' expectations of where the secondary concept should be integrated.
For example, P3 commented that the placement ``\textit{makes more sense,}'' while P5 noted that it ``\textit{follows physical intuition.}''
Our method also shows stable recognition accuracy and response times across blending types.
While baseline methods show a clear drop in performance for positive--negative blending, our method maintains consistent performance, though at a modest cost in response time that reflects the more subtle way concepts are encoded.
As P7 noted, ``\textit{You kind of have to look for it, but once you find it, it’s actually very clear.}'' 

The questionnaire results are shown in Figure~\ref{fig:eva_user}.
\shishi{%
For each metric and blending type, a Friedman test revealed significant differences across the five methods in all cases ($p < 0.001$); post-hoc pairwise two-tailed Wilcoxon signed-rank tests with Holm--Bonferroni correction show our method scores comparably to pixel-based baselines on recognizability (p = 0.18) while significantly outperforming vector-based baselines on recognizability (p = 0.004, 0.006 for Claude Sonnet 4.6 and GPT-5.3, respectively) and outperforming all baselines on creativity and expressiveness.
To understand what drives these differences, we thematically analyze participants' open-ended responses next.
}%

\strike{%

\medskip
\noindent\textbf{Recognizability.}
Our method achieves high scores under both positive ($\bar{x}=6.08$) and negative ($\bar{x}=6.00$) blending, comparable to pixel-based baselines, with no significant difference observed (p = 0.18), while significantly outperforming vector-based methods (p = 0.004 and 0.006 for Claude Sonnet 4.6 and GPT-5.3 respectively). 
Participants noted that while one of the concepts may be less immediately visible, especially in negative blending, where it can appear ``\textit{hidden}'' in the background -- it remains easy to recognize after a brief inspection. This process was often described as enjoyable, as ``\textit{it’s not immediately obvious, but figuring it out is actually enjoyable,}'' (P12). In contrast, more direct baseline results were frequently described as easier to recognize at first glance but “too obvious” or lacking engagement.
Overall, these results suggest that our method maintains high recognizability while introducing a level of perceptual discovery that enhances user engagement, without sacrificing interpretability.

\medskip
\noindent\textbf{Creativity.}
Our method receives the highest creativity ratings under both positive ($\bar{x}=6.42$) and negative ($\bar{x}=6.55$) blending, significantly outperforming all baselines.  
Participants consistently perceived the placement of the secondary concept in our results as more meaningful and surprising. In particular, several participants described the composition as ``\textit{a serendipitous discovery}'' (P3), noting that ``\textit{the placement feels surprisingly natural}'' (P7), even though it was not immediately expected. Others commented that ``\textit{it’s not something I would expect, but it makes a lot of sense once you see it}'' (P11), highlighting the balance between novelty and coherence.  
In contrast, baseline methods were often perceived as less intentional in their spatial composition, where concepts are placed in an expected way without deeper integration. While such designs are easy to understand at first glance, participants noted that they can feel ``\textit{less engaging}'' (P6) or ``\textit{something you can come up with right away}'' (P9).

\medskip
\noindent\textbf{Expressiveness.}
Our method is consistently rated highest in expressiveness under both positive ($\bar{x}=6.35$) and negative ($\bar{x}=6.42$) blending.
Participants frequently attributed these ratings to how the two concepts are visually integrated rather than simply co-existing. In our results, the concepts often share boundaries and form a unified structure, leading to stronger visual clarity. As one participant noted, ``\textit{these concepts seem to reinforce each other, rather than compete for attention}'' (P4). 
Baseline results tended to separate the concepts into distinct objects. Although both concepts ``\textit{are easy to recognizable},'' participants noted that ``\textit{it is not convincing as a blending but feels more like a composition},'' (P1, P7) leading to a weaker overall impression.

}%

\shishi{%

\medskip
\noindent\textbf{Interpreting blending regions.}
Beyond rating the final image, novice participants described a distinctive, layered process of noticing how the two concepts share structure.
As P4 described: ``\textit{At first glance, I only see one concept. But when I look more closely, I notice another one hidden in the image. Then, looking even more carefully, I realize that the second concept is also helping form the first}.''
For example, a region that reads as a leaf or patch of forest on its own also shapes the contour of the primary animal. 
Participants contrasted this with baseline outputs, which they described as feeling like ``\textit{a collage game, where two separate things are pasted together}'' (P8).
In other words, easier to recognize at a glance, but without the two concepts sharing structure.

Participants also connected \textit{where} a concept was placed to how the result was interpreted.
P5 noted that embedding a building in the negative space of an excavator, rather than simply placing it on top, ``\textit{makes the image feel much more ironic}.''
Several participants noted that positive--negative blends were harder to predict than positive--positive ones, but found the model's region choices reasonable in retrospect.
For example, P11 was ``\textit{quite surprised that the model could find suitable negative-space regions}, especially when it chose a location different from the one I had imagined,'' but added that ``\textit{once I saw it, though, the choice felt quite reasonable}.''

\medskip
\noindent\textbf{Emergent higher-order meaning.}
Several participants described results as evoking meaning beyond the two source concepts themselves, rather than simply displaying both.
P6 explained that the penguin--bottle blend was ``\textit{not just a penguin and a bottle}'' but instead evoked ``\textit{plastic pollution and the harm it causes to wildlife}.''
Similarly, P2 described a blend combining a mother-and-child scene with the shape of an ear as ``\textit{very gentle}'' and evocative of ``\textit{a mother softly humming a lullaby while putting her child to sleep}.''
This was not an association either source concept presented in isolation, but emerged from their visual blend using our approach.
Reflecting on this quality more broadly, P10 suggested that ``\textit{this kind of positive--negative space design would work especially well for fables}'', noting that the form is ``\textit{simple, but powerful}.''

These responses suggest that beyond structurally integrating two concepts, our approach can support the kind of emergent, compound meaning central to concept blending in the cognitive-science sense~\cite{Fauconnier:1998:ConceptualBlending}, a quality participants did not attribute to baseline outputs.
We revisit this connection in Section~\ref{sec:design_implications}.

}%

\begin{figure}[!t]
    \centering
    \includegraphics[width=\linewidth]{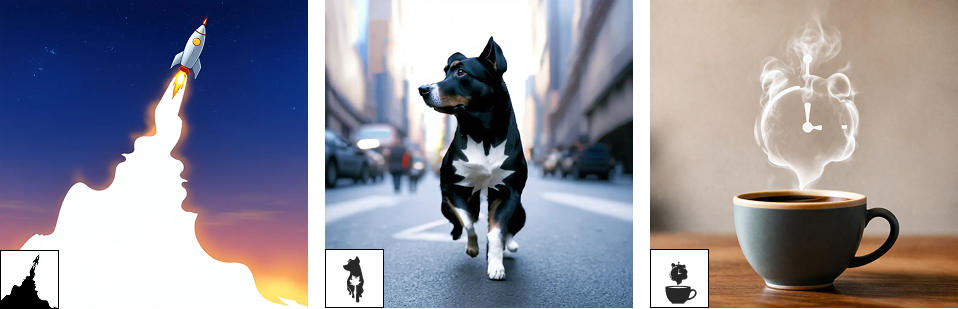}    
    \caption{%
        Case 1: Using the result as control medium for image generation.
    }%
    \label{fig:eva_case_control}
\end{figure}

\begin{figure}[!t]
    \centering
    \includegraphics[width=\linewidth]{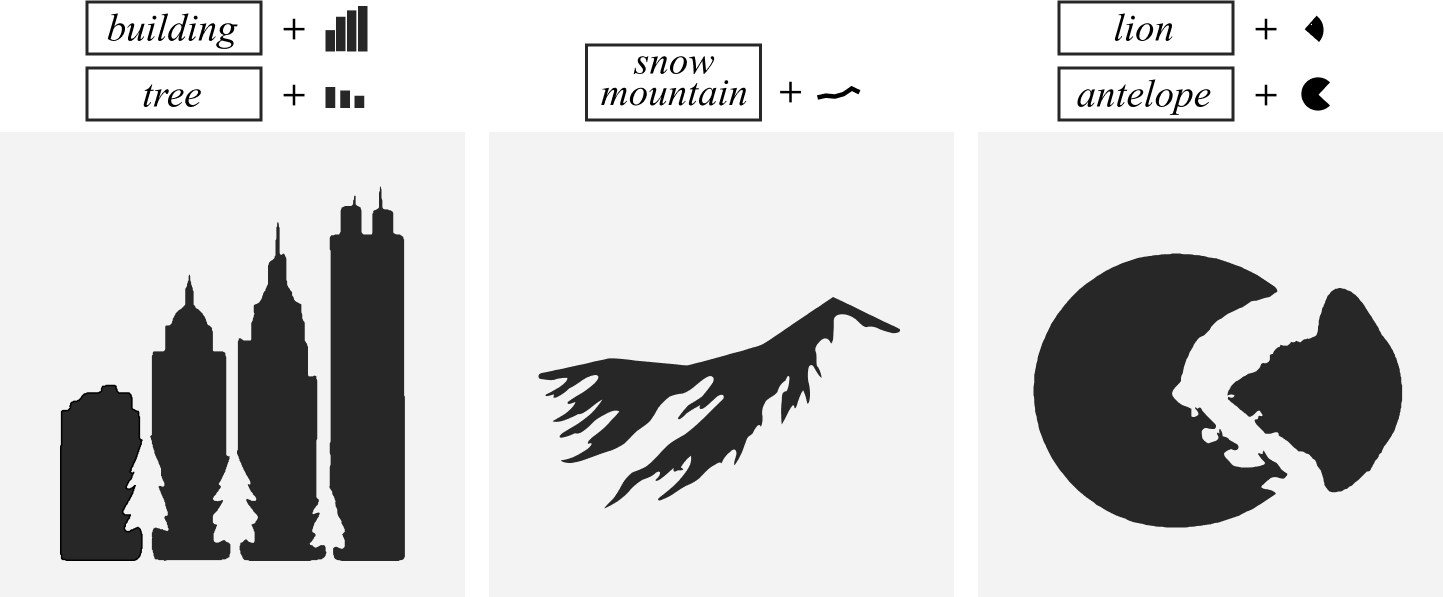}    
    \caption{%
        Case 2: Applying the pipeline for data visualization.
    }%
    \label{fig:eva_case_chart}
\end{figure}

\subsection{Case Study}
\label{sec:case_study}

\shishi{Beyond single-object blending, we finally demonstrate the generalizability of our pipeline in two downstream scenarios that require both structural control and semantic precision: controllable image generation and data-driven infographics.}

\subsubsection{Controllable Image Generation.}
Controllable generative models often rely on external conditioning signals, such as Canny edges or depth maps, to achieve spatial control~\cite{zhang2023adding,tan2025ominicontrol}.
However, obtaining high-quality conditioning inputs is non-trivial and typically requires additional preprocessing.
Our pipeline directly produces clean and structured contour representations that can serve as an effective control medium for downstream generation.

As shown in Figure~\ref{fig:eva_case_control}, we use the generated contour as a Canny condition in ControlNet to achieve precise spatial control in complex image synthesis.
Moreover, the editability of the SVG representation allows direct manipulation of control points, providing more effective editing than raster-based conditioning.
This is particularly beneficial in iterative or sequential design scenarios, where modifying raster conditions can be cumbersome, while tweaking SVG contours is more efficient.


\subsubsection{Infographic-ready visualizations}

Infographics often embed semantic concepts into data charts to enhance expressiveness while preserving quantitative accuracy. 
In our formulation, the chart serves as the \textit{primary concept}, whose shape should be preserved, while additional semantics are introduced as secondary concepts through blending.

Existing approaches for chart blending are largely limited to positive--positive blending~\cite{Wu:2023:Viz2Viz, xiao2026chartist}. 
In contrast, our method explicitly utilizes both positive and negative space, enabling more expressive and structured integrations, as shown in Figure~\ref{fig:eva_case_chart}. 
For example, we transform bar charts into building-like structures via positive blending, while using negative space to form tree silhouettes. 
This enables a single visualization to convey dual data-driven semantics, creating a clear visual contrast between urbanization and the environment.

Moreover, prior methods are predominantly pixel-based, making it difficult to adjust results when they deviate from the underlying data. 
In contrast, our SVG representation supports direct and efficient editing, allowing data-consistent refinements without regenerating the entire image.
Adapting our pipeline to charts requires respecting data encoding constraints. 
Unlike natural shapes, chart elements must preserve key geometric properties. 
To accommodate this, we replace geometry-driven region detection with a human-in-the-loop initialization, where approximate blending regions are specified and further refined by the planner agent.

\section{Discussion}

\shishi{%
Our evaluation results point to concrete implications for how visual blending systems should be designed, while also surfacing open questions about the broader vision of concept-aware generation and the limits of our current system. We discuss each in turn below.
}%

\subsection{Design Implications}
\label{sec:design_implications}

\shishi{%
Our evaluation surfaces two implications for the design of visual blending systems.

First, \textbf{spatial composition should be treated as a first-class design control}, not an implementation detail. Our Task 1 results (Section~\ref{sec:user_study}) show that where a secondary concept is integrated has a direct, measurable effect on how well the result aligns with human expectations (IoU), and our questionnaire results further show that this spatial decision shapes perceived recognizability, creativity, and expressiveness. This suggests that future blending systems, whether pixel- or vector-based, should expose and support explicit reasoning about \textit{where} to blend, rather than treating spatial placement as a byproduct of the generation process.

Second, \textbf{meaning-driven blending requires higher-level concept support}. Our pipeline provides the spatial and contour-level mechanism for integrating two concepts, and our user study suggests this is sufficient to support recognition of emergent, compound meaning that goes beyond simple juxtaposition.
For example, participants read our factory-skull blend as evoking pollution, and our building-tree blend as staging a contrast between urbanization and the environment (Section~\ref{sec:user_study_results}; Figure~\ref{fig:eva_case_chart}), interpretations that were less accessible to participants in the baseline outputs.
This suggests our approach offers one concrete technical path toward what Fauconnier and Turner term ``concept blending''~\cite{Fauconnier:1998:ConceptualBlending}: the emergence of new meaning from two inputs, rather than their side-by-side display.

At the same time, our pipeline does not reason about \textit{which} concept pairs or regions are likely to produce such emergent meaning; this is left to the user's initial framing.
Future systems could build on our spatial mechanism with higher-level support for concept-pair ideation and region selection driven explicitly by intended metaphorical meaning, moving closer to full computational support for concept blending in Fauconnier and Turner's sense.
}%

\subsection{Toward Agents that Do More with Less}
\label{sec:toward_agents}
\shishi{%
Beyond these interface-level implications, our experience building this pipeline points to a broader gap in how generative agents are trained and evaluated.
}%
In the rush to create agents that build ever bigger, brighter, and more complex artifacts, there's a subtle yet massive implication that our work reveals -- models struggle to generate ``less''.
In other words, they often struggle with understanding hidden concepts that are implicit to human viewers.

In our work, one of the most difficult parts of developing our agents was training them to understand negative concepts.
How can we build agents that respect this subtlety?
The ML and HCI community currently lack the tools and infrastructure needed to achieve this vision.
For example, there is a lack of datasets and models that can map negative concepts to design.
Such a dataset could allow agents to train autonomously against new objectives, opening the door for developing foundation models that understand how to judge positive (i.e. explicit) and negative (i.e. implicit) concepts in images.
This would open up several opportunities.
These models can be integrated into planning pipelines as design-aware modules, replacing current VLMs. They can also serve as evaluators to guide and refine agent behavior in collaboration with human designers.
This points to another gap in the community -- there are no common guidelines for what negative concepts are, and how to automatically identify them reliably in graphic design.

This idea extends beyond graphic design. 
In writing, agents are widely used for generating prose, yet a common complaint is that outputs are overly verbose, introducing unintended ideas and directions. 
This mirrors visual blending, where models tend to add more elements rather than convey meaning through structure.
Instead, agents should aim to do more with less, planning around concepts as they generate. 
Analogous to our framework, text generation could distinguish \textit{explicit} content while encoding \textit{implicit} meaning in the "negative" space between sentences. 
Although not a standard training objective, this principle is often regarded as a hallmark of strong writing, and recent work has begun to explore it through goal-oriented reasoning and iterative, human-centered generation.

\subsection{Limitations and Future Work}
\label{sec:limitations}

\shishi{%
Our user study (Section~\ref{sec:user_study}) recruited novice participants to evaluate audience-level perception of outputs, since a successful blend must ultimately communicate to general viewers.
Yet this leaves open how practicing graphic designers would use the system in their own workflow, and whether our design guidelines (\textbf{DG1--DG3}) match professional practice.
Relatedly, our evaluation centers on perception of pipeline outputs rather than interactive use.
Future work should evaluate the interface directly with expert users completing full design tasks.
As discussed in Section~\ref{sec:toward_agents}, the community also lacks datasets and models mapping negative, implicit concepts to design -- a prerequisite for the meaning-driven concept support outlined in Section~\ref{sec:design_implications}.
Finally, extending our two case studies (Section 5.3) to domains like brand identity and logotype design, where negative space and structural metaphor are already core techniques, is a promising direction for future work.
}%

\section{CONCLUSION}


We presented an automatic composition-aware pipeline for visual concept blending that explicitly models where and how concepts are integrated within a shared contour. By combining geometric reasoning, agent-based planning, and hybrid pixel--vector optimization, our approach produces coherent, expressive, and editable vector designs. Our results demonstrate improvements in expressiveness, recognizability, and creativity, while supporting flexible applications across multiple design domains. We hope this work highlights the importance of spatial composition in generative design and inspires future tools for creative visual communication.


\bibliographystyle{ACM-Reference-Format}
\bibliography{main}

\clearpage
\appendix

\section{Input Processing}
In our implementation, we empirically set each closed contour to contain 150 control points. We denote the resulting contour as $\mathcal{P} = {\{p_i\}}_{i=1}^N$, where each $p_i \in \mathbb{R}^2$ is ordered along the boundary.

\section{Stage~II: Blending and Refinement}
In Stage~II, our pipeline integrates a secondary concept into a selected contour of the primary concept using a hybrid pixel–vector strategy that combines pixel-space generation and vector-level optimization.
Here we provide precise mathematical definitions and algorithmic details.

\subsubsection{Concept Injection via Inpainting}

The blending mask is constructed by rasterizing $S$ as a polyline.
To provide sufficient space for generation, we progressively expand the mask by increasing the width of the polyline across inpainting iterations.

\subsubsection{Pixel–Vector Alignment}

To integrate rasterized details into a vector representation, we perform a constrained optimization over the selected segment $S$.
We extract the target contour from the inpainted region as a point cloud $\mathcal{T}$, and rearrange the Bézier control points within $S$ by solving the following optimization problem
\begin{equation}
\min_{S} \ \mathcal{L}_{\text{chamfer}}(S, \mathcal{T}) + \lambda \mathcal{L}_{\text{bending}}(S).
\end{equation}
The Chamfer loss aligns the vector contour with the target geometry \(\mathcal{L}_{\text{chamfer}}(S, \mathcal{T}) = \tfrac{1}{|S|} \sum_{p_i \in S} \min_{q_j \in \mathcal{T}} \|p_i - q_j\|_2^2.\) To ensure structural coherence, we introduce a bending loss that regularizes local curvature:
\(\mathcal{L}_{\text{bending}}(S) = \sum_{p_i \in S} \left(1 - \cos(\theta_i)\right)^2\), where $\theta_i$ is the angle between $(p_i - p_{i-1})$ and $(p_{i+1} - p_i)$.
This term penalizes sharp turns and prevents self-intersections or jagged artifacts during deformation.
Optimizing these objectives yields an updated SVG contour that is geometrically aligned with the generated content while preserving the original parameterization of $\mathcal{P}$.

\subsubsection{Blending Objectives Evaluation}

An evaluator agent assesses the expressiveness of the in-progress blending result.
Formally, we represent the evaluation as a set of concept--score pairs
\(
\mathcal{E} = \{(c_k, s_k)\}_{k=1}^K,  s_k \in [0,1],
\)
where each $c_k$ denotes a semantic component of the blended concept and $s_k$ measures its expressiveness.
At the local level, each component $c_k$ is associated with a contour segment $S_k \subseteq S$, enabling spatially grounded evaluation along the curve. Regions with lower scores indicate insufficient or ambiguous expression of the intended concept.
We use these scores to guide optimization by prioritizing low-expressiveness regions.
In practice, we apply a threshold $\tau = 0.5$ and focus refinement on segments where $s_k < \tau$.

\subsubsection{Contour Optimization}

The SDS objective is defined as
\begin{equation}
\mathcal{L}_{\text{SDS}} = \sum_{k\in [K]} (1 - s_k) \cdot 
\mathbb{E}_{t,\epsilon} \left[
\left\| \epsilon_\theta(x_t; c_k) - \epsilon \right\|_2^2
\right],
\end{equation}
where $x_t$ is the noised rendering of the current contour at timestep $t$, and $\epsilon_\theta$ denotes the noise prediction conditioned on prompt $c_k$.
The loss is weighted by $(1 - s_k)$, prioritizing under-expressed concepts.
We jointly optimize it with $\mathcal{L}_{\text{bending}}$ to avoid contour intersections.

\begin{figure}[t]
    \centering
    \includegraphics[width=\linewidth]{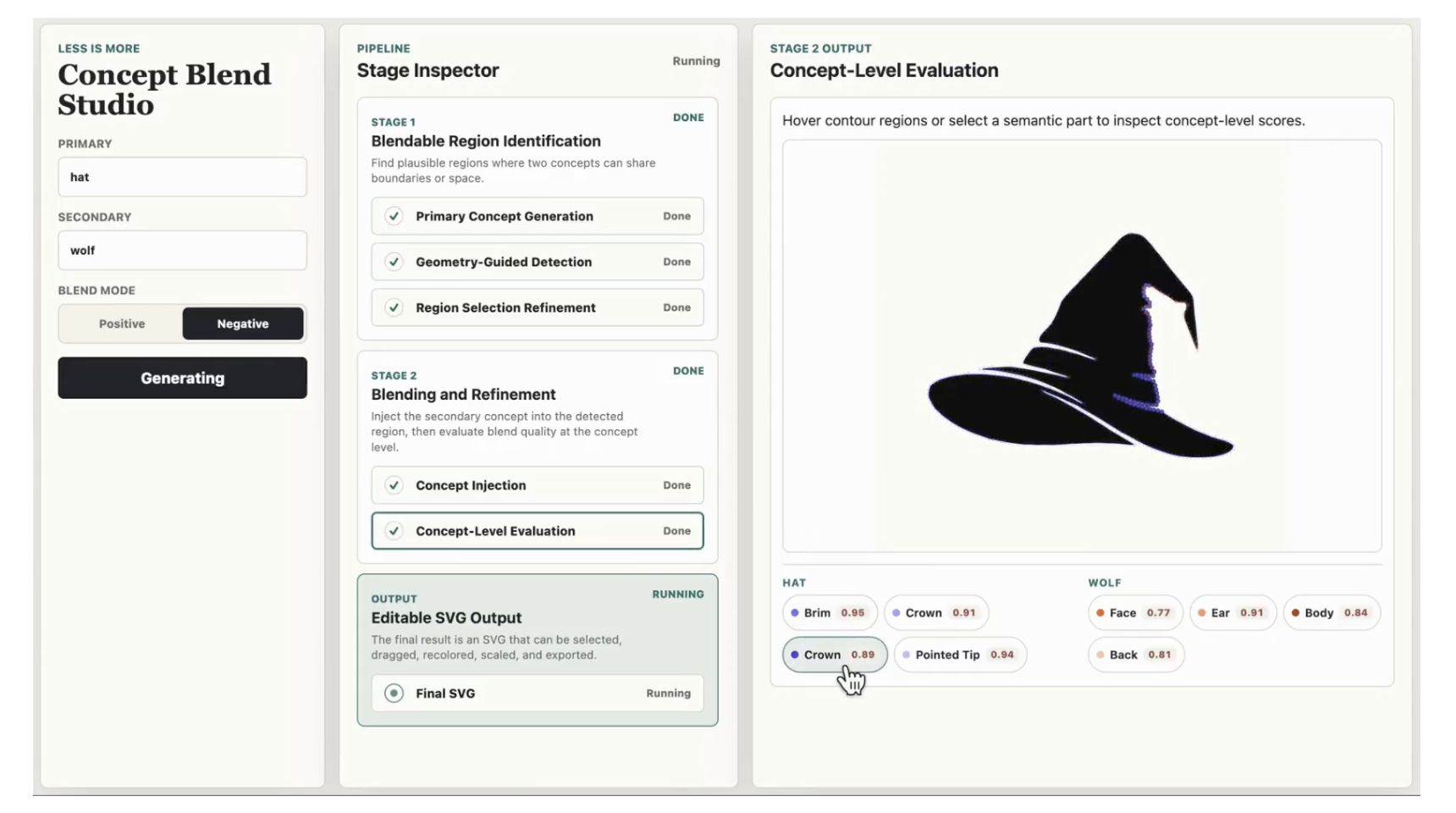}    
    \caption{%
        Screenshot of interface.
    }%
    \label{fig:supp_ui}
    \Description{%
      user interface
    }%
\end{figure}

\section{User Interface Details}
\label{sec:supp-ui}

We developed a prototype interface that exposes the major stages of the visual blending pipeline and allows users to inspect intermediate decisions. As shown in Figure~\ref{fig:supp_ui}, the interface is organized into three panels: input specification, pipeline inspection, and stage-specific output inspection.

The left panel allows users to specify a \textit{primary concept}, a \textit{secondary concept}, and the desired blending mode.
The center panel presents the pipeline as a sequence of inspectable stages.
The right panel visualizes the output associated with the selected pipeline step. 
For instance, during concept-level evaluation, the generated contour is divided into semantic components associated with the primary and secondary concepts. Users can hover over a contour region or select a semantic component to inspect its evaluation score.
The final output stage exposes the generated SVG as an editable artifact. Because the output remains in vector format, users can subsequently select, move, recolor, scale, or export individual components.


\end{document}